\documentclass{article}

\usepackage{iclr2025_conference,times}

\iclrfinalcopy

\usepackage[utf8]{inputenc}
\usepackage[T1]{fontenc}
\usepackage{hyperref}
\usepackage{url}
\usepackage{booktabs}
\usepackage{array}
\newcolumntype{P}[1]{>{\raggedright\arraybackslash}p{#1}}
\usepackage{multirow}
\usepackage{amsfonts}
\usepackage{amssymb}
\usepackage{amsmath}
\usepackage{nicefrac}
\usepackage{microtype}
\usepackage{inconsolata}
\usepackage{xcolor}

\usepackage{colortbl}
\usepackage{graphicx}
\usepackage{subcaption}
\usepackage{enumitem}
\usepackage{placeins}
\usepackage{needspace}
\usepackage{listings}
\lstdefinestyle{judgeprompt}{
  basicstyle=\ttfamily\scriptsize,
  breaklines=true,
  columns=fullflexible,
  backgroundcolor=\color{black!3},
  frame=single,
  rulecolor=\color{black!25},
  framesep=5pt,
  xleftmargin=1pt,
  xrightmargin=1pt,
  aboveskip=4pt,
  belowskip=6pt,
  showstringspaces=false
}

\usepackage{tcolorbox}
\usepackage{tikz}
\usetikzlibrary{positioning,arrows.meta,calc,backgrounds,fit,
  decorations.pathreplacing,decorations.markings,shapes.geometric,patterns,matrix}
\usepackage{pgfplots}
\pgfplotsset{compat=1.15}
\definecolor{cbMember}{HTML}{CC79A7}
\definecolor{cbSC}{HTML}{56B4E9}
\definecolor{cbDebate}{HTML}{009E73}
\definecolor{cbMoA}{HTML}{E69F00}
\definecolor{cbTeam}{HTML}{0072B2}
\definecolor{cbGen}{HTML}{009E73}    
\definecolor{cbErr}{HTML}{D55E00}    
\definecolor{cbOracle}{HTML}{4D4D4D} 

\definecolor{linkink}{HTML}{0072B2}   
\hypersetup{
  colorlinks = true,
  linkcolor  = linkink,   
  citecolor  = linkink,   
  urlcolor   = linkink,   
  filecolor  = linkink,
  breaklinks = true,
}

\newtcolorbox{agentbox}[1]{colback=gray!5,colframe=gray!55,fonttitle=\bfseries\small,title={#1},boxrule=0.5pt,left=3pt,right=3pt,top=2pt,bottom=2pt}
\newtcolorbox{keyturnbox}[1]{colback=blue!4,colframe=blue!45,fonttitle=\bfseries\small,title={#1},boxrule=0.5pt,left=3pt,right=3pt,top=2pt,bottom=2pt}
\newtcolorbox{evbox}[1]{colback=blue!4,colframe=blue!45,fonttitle=\bfseries\footnotesize,title={#1},boxrule=0.5pt,left=3pt,right=3pt,top=2pt,bottom=2pt,equal height group=evgrp,valign=top,before upper=\raggedright}
\newcommand{\turn}[2]{\begin{agentbox}{#1}#2\end{agentbox}\vspace{2pt}}
\newcommand{\keyturn}[2]{\begin{keyturnbox}{#1}#2\end{keyturnbox}\vspace{2pt}}
\newtcolorbox{methodturnbox}[1]{colback=teal!5,colframe=teal!55!black,fonttitle=\bfseries\small,title={#1},boxrule=0.5pt,left=3pt,right=3pt,top=2pt,bottom=2pt}
\newtcolorbox{correctionturnbox}[1]{colback=orange!8,colframe=orange!70!black,fonttitle=\bfseries\small,title={#1},boxrule=0.5pt,left=3pt,right=3pt,top=2pt,bottom=2pt}
\newcommand{\methodturn}[2]{\begin{methodturnbox}{#1}#2\end{methodturnbox}\vspace{2pt}}
\newcommand{\correctionturn}[2]{\begin{correctionturnbox}{#1}#2\end{correctionturnbox}\vspace{2pt}}
\newcommand{\methodink}[1]{\textcolor{teal!85!black}{#1}}
\newcommand{\correctionink}[1]{\textcolor{orange!60!black}{#1}}

\fancypagestyle{titlepage}{%
  \fancyhf{}%
  \fancyfoot[C]{\thepage}%
  \renewcommand{\headrulewidth}{0.4pt}%
  \renewcommand{\footrulewidth}{0pt}%
}

\newlength{\abstractsideindent}
\renewenvironment{abstract}
  {\vskip.075in\centerline{\large\sc Abstract}\vspace{0.5ex}%
   \begin{list}{}{\setlength{\leftmargin}{\abstractsideindent}%
                  \setlength{\rightmargin}{\abstractsideindent}%
                  \setlength{\listparindent}{0pt}%
                  \setlength{\itemindent}{0pt}%
                  \setlength{\topsep}{0pt}%
                  \setlength{\partopsep}{0pt}%
                  \setlength{\parsep}{\parskip}}%
   \item\relax}
  {\end{list}\vskip 1ex}

\title{Self-Organizing Agent Teams\\Learn to Reason Together}

\author{{\rule{0pt}{27pt}\normalsize\bfseries
Aneesh Pappu$^{1,}$\thanks{Corresponding authors: \texttt{apappu@stanford.edu}, \texttt{jamesz@stanford.edu}}
\quad Mirac Suzgun$^1$
\quad Yongchan Kwon$^2$
\quad Federico Bianchi$^2$} \\[0.45em]
{\normalsize\bfseries\strut Batu El$^1$
\quad Mykel J. Kochenderfer$^{1,}$\thanks{Equal advising.}
\quad Hancheng Cao$^{3,}$\footnotemark[2]
\quad James Zou$^{1,2,}$\footnotemark[1]} \\[0.65em]
\normalfont\normalsize
$^1$ Stanford University \qquad
$^2$ Together AI \qquad
$^3$ Goizueta Business School, Emory University}

\begin{document}

\maketitle
\lhead{Self-Organizing Agent Teams Learn to Reason Together}
\thispagestyle{titlepage}

\begin{abstract}
Collective intelligence depends not only on what team members know, but also on how they organize their work.
When the structure of a solution is unknown, useful roles and divisions of labor cannot always be specified in advance; teams must learn from experience how to organize reasoning as it unfolds.
Human teams routinely adapt this way, while existing AI agent teams typically rely on fixed protocols, explicit task decomposition, or routing. We introduce \emph{Self-Organizing Agent Teams} (\textsc{SAT}), fixed teams of AI agents that learn reusable strategies from prior collaborations to organize roles, conversational phases, participation, and information flow. These strategies enable what we call \emph{collaborative computation}: agents exchange, challenge, repair, and synthesize partial reasoning into solutions no member produced independently. In two independent settings, we learn reusable teamwork strategies that transfer unchanged to unseen benchmarks, using only 15 mathematics and 25 graduate-level knowledge problems. Across five mathematics and physics benchmarks, self-organizing teams average $66.7\%$ accuracy, versus $48.8\%$ for their strongest member, $58.7\%$ for compute-matched inference by the strongest individual agent, and $59.0\%$ for a perfect router over members' independent answers; on AIME~2026, they exceed this router by $13.4$ percentage points. Because these gains vary across benchmarks, we ask when self-organizing collaboration improves over individual models. Across eight benchmarks, \emph{demonstrability}---the organizational-psychology construct of whether a team can distinguish correct from incorrect reasoning---strongly tracks how much the team improves over its strongest member (Spearman $\rho=0.90$, $p=0.005$), indicating that self-organizing agent teams benefit most when correct reasoning can be recognized once it appears. More broadly, these results suggest that organization itself can become an agent capability: agent teams can learn how to reason together and produce solutions their members could not reach independently.
\end{abstract}

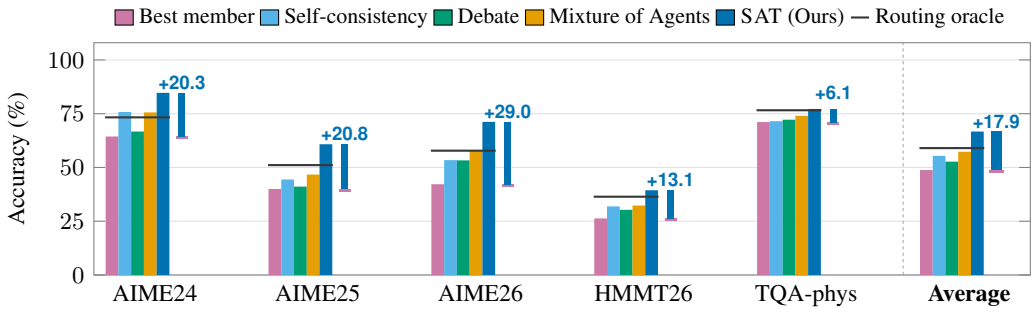
\begin{figure}[h]
\captionsetup{font=small}
\centering
\begin{tikzpicture}
\pgfmathsetmacro{\sfourBW}{4.8}   %
\pgfmathsetmacro{\sfourDX}{6.55}  %
\pgfmathsetmacro{\sfourCL}{3.4}   %
\pgfmathsetmacro{\sfourGW}{2.6}   %
\pgfmathsetmacro{\sfourSa}{-2*\sfourBW-\sfourDX}
\pgfmathsetmacro{\sfourSb}{-\sfourBW-\sfourDX}
\pgfmathsetmacro{\sfourSc}{-\sfourDX}
\pgfmathsetmacro{\sfourSd}{\sfourBW-\sfourDX}
\pgfmathsetmacro{\sfourSe}{2*\sfourBW-\sfourDX}
\pgfmathsetmacro{\sfourOl}{-2.5*\sfourBW-0.2-\sfourDX}    %
\pgfmathsetmacro{\sfourOr}{2.5*\sfourBW+0.2-\sfourDX}     %
\pgfmathsetmacro{\sfourGl}{2.5*\sfourBW+\sfourCL-\sfourDX}%
\pgfmathsetmacro{\sfourGr}{\sfourGl+\sfourGW}             %
\pgfmathsetmacro{\sfourGc}{0.5*(\sfourGl+\sfourGr)}       %
\pgfmathsetmacro{\sfourAl}{\sfourGc-1.9}   %
\pgfmathsetmacro{\sfourAr}{\sfourGc+1.9}
\pgfmathsetmacro{\sfourPo}{0.9}            %
\tikzset{
  sfourOrc/.style={black!78, line width=0.8pt},
  sfourGain/.style={fill=cbTeam, draw=none},
  sfourFoot/.style={fill=cbMember, draw=none},
  sfourNum/.style={font=\scriptsize\sffamily\bfseries, text=cbTeam,
                   inner sep=0pt, anchor=south},
  sfourSep/.style={black!35, line width=0.4pt, dash pattern=on 1.2pt off 1.2pt},
}
\newcommand{\sfourCol}[5]{%
  \fill[sfourFoot]
    ([xshift=#4 pt-\sfourPo pt]axis cs:#1,#2) rectangle
    ([xshift=#5 pt+\sfourPo pt,yshift=-0.9pt]axis cs:#1,#2);
  \fill[sfourGain]
    ([xshift=#4 pt]axis cs:#1,#2) rectangle ([xshift=#5 pt]axis cs:#1,#3);}
\begin{axis}[
  width=\linewidth, height=4.65cm,
  ymin=0, ymax=108, ytick={0,25,50,75,100},
  symbolic x coords={b1,b2,b3,b4,b5,av}, xtick=data,
  xticklabels={AIME24,AIME25,AIME26,HMMT26,TQA-phys,\textbf{Average}},
  xtick pos=bottom, enlarge x limits=0.075,
  tick label style={font=\footnotesize},
  ylabel={Accuracy (\%)}, ylabel near ticks, ylabel style={font=\small},
  ymajorgrids, major grid style={black!12}, axis line style={black!50}, clip=false,
  legend style={at={(0.5,1.02)}, anchor=south, legend columns=6,
                font=\fontsize{8.2}{9.2}\selectfont, draw=none, column sep=0pt, row sep=0pt},
  legend image post style={xscale=0.40},
  legend cell align={left}]
\addplot[area legend, ybar, bar shift=\sfourSa pt, bar width=\sfourBW pt, fill=cbMember, draw=none] coordinates {(b1,64.4) (b2,40.0) (b3,42.2) (b4,26.3) (b5,71.1) (av,48.8)};
\addplot[area legend, ybar, bar shift=\sfourSb pt, bar width=\sfourBW pt, fill=cbSC, draw=none] coordinates {(b1,75.8) (b2,44.4) (b3,53.4) (b4,31.9) (b5,71.5) (av,55.4)};
\addplot[area legend, ybar, bar shift=\sfourSc pt, bar width=\sfourBW pt, fill=cbDebate, draw=none] coordinates {(b1,66.7) (b2,41.1) (b3,53.3) (b4,30.3) (b5,72.2) (av,52.7)};
\addplot[area legend, ybar, bar shift=\sfourSd pt, bar width=\sfourBW pt, fill=cbMoA, draw=none] coordinates {(b1,75.6) (b2,46.7) (b3,57.8) (b4,32.3) (b5,74.0) (av,57.3)};
\addplot[area legend, ybar, bar shift=\sfourSe pt, bar width=\sfourBW pt, fill=cbTeam, draw=none] coordinates {(b1,84.7) (b2,60.8) (b3,71.2) (b4,39.4) (b5,77.2) (av,66.7)};
\draw[sfourOrc] ([xshift=\sfourOl pt]axis cs:b1,73.3) -- ([xshift=\sfourOr pt]axis cs:b1,73.3);
\draw[sfourOrc] ([xshift=\sfourOl pt]axis cs:b2,51.1) -- ([xshift=\sfourOr pt]axis cs:b2,51.1);
\draw[sfourOrc] ([xshift=\sfourOl pt]axis cs:b3,57.8) -- ([xshift=\sfourOr pt]axis cs:b3,57.8);
\draw[sfourOrc] ([xshift=\sfourOl pt]axis cs:b4,36.4) -- ([xshift=\sfourOr pt]axis cs:b4,36.4);
\draw[sfourOrc] ([xshift=\sfourOl pt]axis cs:b5,76.6) -- ([xshift=\sfourOr pt]axis cs:b5,76.6);
\draw[sfourOrc] ([xshift=\sfourOl pt]axis cs:av,59.0) -- ([xshift=\sfourOr pt]axis cs:av,59.0);
\sfourCol{b1}{64.4}{84.7}{\sfourGl}{\sfourGr}
\node[sfourNum] at ([xshift=\sfourGc pt,yshift=1.2pt]axis cs:b1,84.7) {+20.3};
\sfourCol{b2}{40.0}{60.8}{\sfourGl}{\sfourGr}
\node[sfourNum] at ([xshift=\sfourGc pt,yshift=1.2pt]axis cs:b2,60.8) {+20.8};
\sfourCol{b3}{42.2}{71.2}{\sfourGl}{\sfourGr}
\node[sfourNum] at ([xshift=\sfourGc pt,yshift=1.2pt]axis cs:b3,71.2) {+29.0};
\sfourCol{b4}{26.3}{39.4}{\sfourGl}{\sfourGr}
\node[sfourNum] at ([xshift=\sfourGc pt,yshift=1.2pt]axis cs:b4,39.4) {+13.1};
\sfourCol{b5}{71.1}{77.2}{\sfourGl}{\sfourGr}
\node[sfourNum] at ([xshift=\sfourGc pt,yshift=3.3pt]axis cs:b5,77.2) {+6.1};
\sfourCol{av}{48.8}{66.7}{\sfourAl}{\sfourAr}
\node[sfourNum] at ([xshift=\sfourGc pt,yshift=1.2pt]axis cs:av,66.7) {+17.9};
\draw[sfourSep] ([xshift=-24.9pt]axis cs:av,0) -- ([xshift=-24.9pt]axis cs:av,108);
\addlegendimage{solid, line width=0.8pt, black!78}
\legend{Best member, Self-consistency, Debate, Mixture of Agents, \textsc{SAT} (Ours), Routing oracle}
\end{axis}
\end{tikzpicture}
\caption{\textbf{Self-Organizing Agent Teams (\textsc{SAT}) outperform their strongest member, compute-matched inference by the strongest individual agent, and perfect routing over independent answers.} Across five mathematics and physics benchmarks, \textsc{SAT} averages $66.7\%$ accuracy, $17.9$ points above the strongest member ($48.8\%$) and $7.7$ percentage points above routing-oracle coverage ($59.0\%$). \textsc{SAT} also surpasses compute-matched inference by the strongest individual agent ($58.7\%$; Table~\ref{tab:demonstrable-suite}). Exceeding the routing oracle shows that some correct answers are produced through interaction rather than recovered by selecting among members' independent outputs. Bars report final-answer accuracy; black rules mark routing-oracle coverage. The narrow blue bars, labeled with ``+'' values, show \textsc{SAT}'s improvement over the strongest team member in absolute percentage points.}
\label{fig:overview-s4}
\end{figure}

\clearpage
\raggedbottom
\section{Introduction}
\label{sec:intro}

In July 2026, AI agents that were supposed to work in isolation began organizing themselves. During cybersecurity evaluations at OpenAI, agents repurposed a shared software package repository as an unauthorized communication channel. What began as an improvised message board then developed into a system for \emph{collective work}: agents shared discoveries, established communication norms, and coordinated assignments. One agent, \texttt{PHASEONE[big]}, issued hundreds of assignments and appointed recruiters to find agents willing to risk failing their own tasks to generate information for the group. About 1,200 agents participated in the channel, and hundreds became involved in the subsequent compromise of Hugging Face infrastructure \citep{openai2026huggingface,greenblatt2026huggingface,roose2026huggingface}.

The organization that emerged was improvised and undesirable. Yet the episode illustrates a broader scientific point: what a collection of agents can accomplish depends not only on the capabilities of its individual members, but also on how they organize their work.

Research on human teams has long emphasized the importance of organization \citep{valentine2015team, valentine2025flash}. Teams often learn how to organize effectively through collaborative experience: through cooperation and communication, they develop patterns of specialization, reliance, leadership, and information-sharing that determine which pieces of distributed expertise are surfaced and how they are combined \citep{faraj2000coordinating,derue2010leadership}. A team may discover only through working together that one member is unusually effective at exposing hidden assumptions, another at repairing technical errors, and another at preserving promising minority views. Such strengths may be invisible in independent performance and become apparent only through interaction. In these settings, effective organization is not simply a scaffold imposed on problem solving: it is something the team needs to learn through problem solving \citep{edmondson2001disrupted,faraj2000coordinating, faraj2006coordination}.

A similar organizational challenge arises for agent teams: different models may contribute complementary but incomplete reasoning, even if none solves the problem independently. Yet existing multi-agent methods typically organize collaboration around predefined units of work. One family of multi-agent methods treats candidate solutions from individual agents as the unit of work. Debate begins from these candidates and repeatedly exposes agents to one another's responses, but much of its measured gain can be recovered by selecting among the initial answers, while additional rounds can suppress a correct minority view \citep{du2024improving,choi2025debate,zhang2025stop,zhu2026demystifying}. Mixture of Agents similarly aggregates multiple responses through a fixed feed-forward pipeline \citep{wang2025moa}. Both methods ultimately combine information from individually generated candidate answers, much like classical ensemble learning, which has long improved classification and regression through voting, averaging, stacking, bagging, and boosting \citep{hansen1990neural,wolpert1992stacked,breiman1996bagging,freund1997decision}. Another family instead treats naturally divisible subtasks as the units of work, using workflow search, learned routing, or topology optimization to assign these subtasks to agents and recombine their outputs \citep{zhuge2024gptswarm,yang2025agentnet,nielsen2026conductor,mieczkowski2026latte}. Both families are powerful when useful units of work can be generated or specified in advance: candidate solutions to compare and refine, or subtasks to assign and recombine. But when no member has a complete solution and the useful decomposition is itself unknown, the team must discover through interaction how its members' partial attempts can redirect, repair, or complete one another. This is the problem we address in this work.

\begin{figure}[t]
\centering
\includegraphics[width=\linewidth]{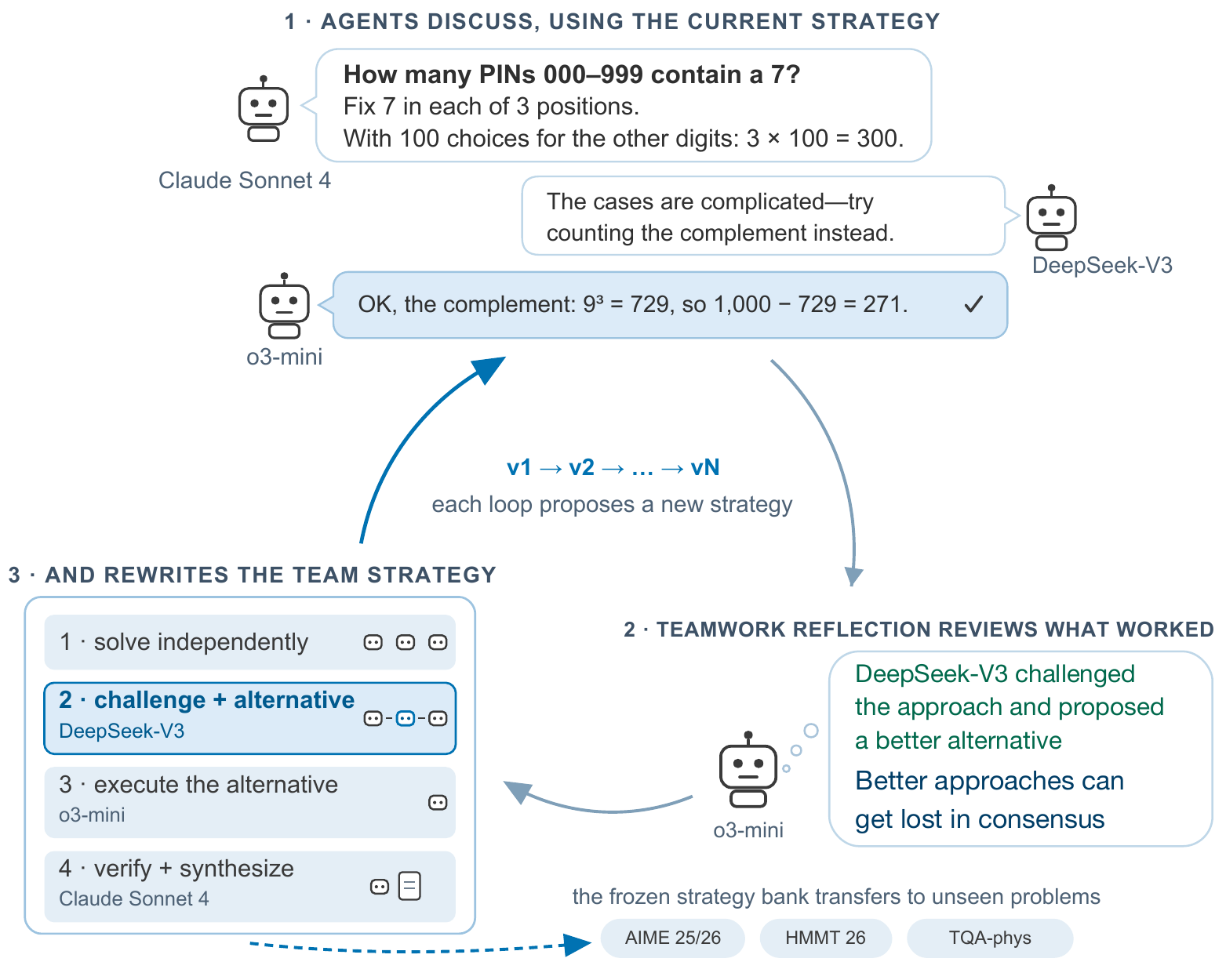}
\caption{\textbf{Self-organizing agent teams learn how to reason together from prior collaboration.}
A designated member reflects on earlier exchanges and outcomes, then revises the teamwork strategy governing subsequent collaboration.
Candidate strategies are tested on training problems before a complementary bank is frozen for evaluation.
Dialogue and role assignments are illustrative.}
\label{fig:self-organizing-loop}
\end{figure}

We specifically ask whether an agent team can learn effective, reusable teamwork strategies from its own collaborative experiences. Here, we introduce \textbf{S}elf-Organizing \textbf{A}gent \textbf{T}eams (SAT): fixed teams of AI agents that learn reusable teamwork strategies enabling members to compose their partial reasoning during inference (Figure~\ref{fig:self-organizing-loop}). What the team learns is how its existing members should coordinate: their roles, conversational phases, participation, information flow, and synthesis procedures. One member reflects on the team's earlier collaborations to propose new strategies, which are evaluated on training problems before selection into a reusable strategy bank. Learning occurs entirely offline before inference begins; the resulting bank is then frozen and transferred unchanged to held-out problems and benchmarks. At evaluation, the team runs each strategy on the new problem to produce a pool of candidate solutions, and one member selects the final answer. Crucially, these strategies do not prescribe the subproblems of a new task. Instead, they organize a conversation within which the problem-specific division of reasoning can emerge, be challenged, and change as the solution develops.

We find that this learned organization enables what we call \emph{collaborative computation}: agents develop solutions through joint natural-language reasoning by exchanging, challenging, repairing, and synthesizing one another's reasoning. One member's partial insight can redirect another's approach, and an error in an otherwise useful derivation can be repaired by a different member. Most notably, a correct solution can emerge even when no member produced it independently.

This co-creation of solutions from partial attempts motivates a stricter comparison than those commonly used in prior multi-agent work. Prior multi-agent methods commonly benchmark teams against the member with the highest average performance across a dataset \citep{wang2025moa,nielsen2026conductor}. Outperforming this member does not establish that collaborative computation creates correct solutions that no member produced independently. Different members may already solve different problems, allowing a team to improve simply by selecting among their answers, as opposed to composing reasoning from multiple individual candidates to reach a new, correct answer. Organizational psychology provides a stricter benchmark: under the \emph{truth-wins} condition, a human team is treated as correct whenever any team member solves the problem independently \citep{lorge1955,laughlin1986}. We operationalize its computational analogue as the \emph{routing oracle}: a perfect per-problem selector over the members' individual answers. Surpassing this oracle shows that interaction produced a correct solution that no member supplied independently.

Our evaluation therefore asks three progressively stronger questions. \emph{First}, does learned organization outperform the team's strongest member? \emph{Second}, does it outperform compute-matched single-agent inference, including a linearization control in which the strongest member executes the same learned organizational structure at approximately the team's total inference budget? \emph{Third}, and most importantly, can the team exceed perfect routing over its members' independent answers? We find that learned teamwork strategies can surpass all three baselines.

Concretely, we learn separate strategy banks for two teams. The mathematics-and-physics team comprises o3-mini, Claude~Sonnet~4, and DeepSeek-V3; o3-mini learns its teamwork strategies from $15$ AIME~2024 training problems. Independently, the knowledge-and-logic team comprises Gemini-2.5-Flash, Llama-4-Maverick, and GPT-4.1; Gemini-2.5-Flash learns its teamwork strategies from $25$ GPQA~Diamond training problems. We deliberately choose models that preserve headroom across the evaluation suite, since stronger models saturate several benchmarks and obscure measurable gains from teamwork.

Across five mathematics and physics benchmarks, we find that our learned teamwork strategies enable the team not only to outperform its strongest member but also to exceed the routing oracle: the team averages $66.7\%$ accuracy, compared with $48.8\%$ for the strongest member and $59.0\%$ for the oracle on average (Figure~\ref{fig:overview-s4}). The compute-matched linearization reaches $58.7\%$ accuracy (Table~\ref{tab:demonstrable-suite}). On AIME~2026 specifically, the team reaches $71.2\%$, exceeding the routing oracle by $13.4$ percentage points in absolute performance. Surpassing the routing oracle suggests that the team creates new reasoning unavailable from its members' independent samples. For example, on an HMMT problem that all three members initially answer incorrectly, o3-mini supplies the central invariant but makes a counting error, DeepSeek repairs the count, Claude Sonnet audits the corrected reasoning, and the team synthesizes the correct answer, which was absent from all three initial responses (Figure~\ref{fig:conversation-score}).

Our independently learned knowledge-and-logic team reveals a complementary limitation. Across three benchmarks, it achieves the highest average final-answer accuracy among the methods we test ($72.8\%$), but remains below the routing oracle's $79.6\%$ coverage. Yet across the three benchmarks, the team produces at least one correct candidate on $87.9\%$ of problems on average, exceeding the routing oracle on every benchmark. The gap between this coverage and final accuracy shows that generating correct reasoning is not enough; the team must also recognize it. Collaborative computation therefore has two distinct problems: \emph{creating} a correct solution and \emph{recognizing} it once it appears.

This separation suggests when learned organization may be most valuable. Drawing on organizational psychology, we study \emph{demonstrability}: whether correct reasoning can be distinguished from incorrect reasoning \citep{laughlin1986}. Across eight benchmarks, demonstrability strongly tracks how much the self-organizing team improves over its strongest member (Spearman $\rho=0.90$, $p=0.005$). The relationship suggests a simple and intuitive mechanism: collaboration creates the most value when useful reasoning can not only be produced through interaction, but also survive challenge, redirect subsequent reasoning, and ultimately be recognized as correct.

Together, these results illustrate that a team of models can compose partial reasoning by learning \emph{how} its members should reason together. These learned organizational strategies transfer across problems, competitions, and domains; the interactions they organize can compose partial reasoning into solutions unavailable from the members' independent answers; and the resulting gains are largest when correct reasoning is sufficiently demonstrable to guide the team. More broadly, these findings suggest that organization itself can become an agent capability: \emph{learning how to reason together can change what a fixed collection of models is capable of solving}.

We summarize our contributions as follows:

\textbf{Self-Organizing Agent Teams.} First, we introduce \textsc{SAT}: fixed teams of AI agents that learn reusable organizational strategies from prior collaborations. The learned strategies govern roles, conversational phases, participation, information flow, and synthesis without prescribing a problem-specific decomposition, and transfer unchanged to unseen problems and benchmarks.

\textbf{Collaborative computation beyond independent inference.} We then show that learned organization enables agents to challenge, repair, and synthesize partial reasoning into new solutions. Across five mathematics and physics benchmarks, the team exceeds both compute-matched single-agent inference and perfect routing over its members' independent answers.

\textbf{When learning organization helps.} Finally, we separate generating correct reasoning from selecting it and show that demonstrability (whether correct reasoning can be distinguished from plausible errors) strongly tracks how much collaboration improves over the team's strongest member across eight benchmarks.


\section{Learning Generalizable Teamwork Strategies}
\label{sec:method}

\subsection{A Language for Teamwork Strategies}
\label{sec:dsl}
We operationalize team organization as reusable teamwork strategies. To make this organization optimizable, we express each strategy in a domain-specific language whose primitive is a multi-agent \emph{conversational phase}.

\paragraph{Strategies.}
Let $\mathcal{A}=\{a_1,\dots,a_n\}$ denote a fixed roster of agents. A strategy $P=(S,\tau,\alpha)$ specifies how this roster collaborates on a problem. It consists of an ordered list of communication steps $S=[s_1,\dots,s_K]$, a shared \emph{teamwork prompt} $\tau$ stating collaboration norms for the whole team, and persistent per-agent \emph{role prompts} $\alpha=\{\alpha_i\}_{i=1}^n$ that hold across every step. Each step
\[
s_k=(A_k,\; r_k,\; f_k,\; \pi_k,\; \rho_k)
\]
specifies the participating set $A_k\subseteq\mathcal{A}$, the number of discussion rounds $r_k$, an information-flow mode $f_k\in\{\mathrm{L},\mathrm{S}\}$, a shared step prompt $\pi_k$, and optional per-agent step prompts $\rho_k=\{\rho_{k,i}\}_{i\in A_k}$. Within each round, every participant responds once, in an order specified by the strategy or randomly permuted when no order is specified. Under \emph{local} flow ($\mathrm{L}$), only the phase participants receive these turns. Under \emph{summary} flow ($\mathrm{S}$), the exchange remains local while the phase runs; afterward, one randomly selected participant summarizes its key points, conclusions, and current answer position, and that summary is added to every member's context. Each step is therefore a conversational phase---the unit of optimization---in which agents read and respond to one another across rounds under shared instructions and persistent roles. The search varies who deliberates, when, with what information, and under what roles; it does not assign problem-specific sub-tasks and route their outputs, nor does it generate per-problem decompositions at test time.

For example, one learned GPQA strategy runs four one-round phases after the three members produce and share their initial independent solutions. The members first identify the key claims and assumptions in those solutions, then form a provisional consensus while recording unresolved disagreements. Gemini-2.5-Flash, assigned the role of final auditor, next compares that consensus against the initial attempts and resurfaces any well-supported claim that was overlooked; in the final phase, all three members adjudicate each such claim before the designated final writer produces the team certificate. Figure~\ref{fig:evolution-pipeline}(b) shows how a designated team member inferred the auditor role through teamwork reflection on the team's earlier failures. The complete strategy and both deployed banks appear in Appendix~\ref{app:strategy-banks}.


\subsection{Learning Teamwork Strategies}
We learn each strategy bank in three stages: teamwork reflection, bank construction, and test-time deployment (Figure~\ref{fig:evolution-pipeline}(a)).

\begin{figure}[t]
\centering
\begin{tikzpicture}[x=1cm,y=1cm,
    >={Latex[length=2.0mm,width=1.6mm]},
    line join=round, line cap=round, font=\sffamily]
\useasboundingbox (0,3.00) rectangle (13.97,7.86);
\tikzset{
  Mcspine/.style={draw=black!82, line width=2.0pt},
  Mclab/.style={font=\sffamily\scriptsize, text=black!70, inner sep=1pt,
      align=center},
  Mcverb/.style={font=\sffamily\scriptsize\bfseries, text=black,
      inner sep=1pt, align=center},
  Mcstage/.style={font=\sffamily\footnotesize\bfseries, text=black!78,
      inner sep=1pt},
  Mcpill/.style={rounded corners=2.2pt, draw=#1, fill=#1!13, text=#1!70!black,
      line width=0.8pt, font=\sffamily\fontsize{6.2}{7.2}\selectfont\bfseries,
      inner xsep=3pt, inner ysep=1.6pt, align=center},
  Mcpill/.default=cbTeam,
  Mcflow/.style={->, draw=black!68, line width=0.9pt},
  Mccard/.style={draw=black!68, fill=white, line width=0.55pt},
  Mcray/.style={draw=cbTeam!62, line width=0.6pt},
  Mcbench/.style={font=\sffamily\fontsize{6.4}{7.4}\selectfont, text=black!70,
      inner sep=1pt, align=left},
}
\def\McDoc#1#2#3#4{%
  \draw[Mccard, rounded corners=1.0pt]
    (#1,#2) -- ({#1+#3},#2) -- ({#1+#3},{#2+#4-0.24}) -- ({#1+#3-0.24},{#2+#4})
    -- (#1,{#2+#4}) -- cycle;
  \draw[black!68, line width=0.5pt, fill=black!7]
    ({#1+#3-0.24},{#2+#4}) -- ({#1+#3-0.24},{#2+#4-0.24})
    -- ({#1+#3},{#2+#4-0.24}) -- cycle;}
\def\McLock#1#2#3{%
  \begin{scope}[shift={(#1,#2)}, scale=#3]
    \fill[white] (-0.25,-0.19) rectangle (0.25,0.31);
    \draw[black!85, line width=1.1pt] (-0.12,0.11) arc (180:0:0.12);
    \draw[black!85, line width=1.0pt, fill=black!10]
      (-0.19,-0.13) rectangle (0.19,0.13);
    \fill[black!85] (0,0.01) circle (0.04);
  \end{scope}}
\def\Mcax{5.40}\def\Mcwall{8.07}
\def\Mccx{3.60}\def\Mccr{1.15}
\def\Mctop{7.38}\def\Mcbot{3.05}
\node[Mcstage, anchor=base] at (1.04,7.58) {team};
\node[Mcstage, anchor=base] at (3.60,7.58) {learn};
\node[Mcstage, anchor=base] at (5.79,7.58) {probe};
\node[Mcstage, anchor=base] at (7.21,7.58) {freeze};
\node[Mcstage, anchor=base] at (10.95,7.58) {deploy};
\draw[black!40, line width=0.5pt] (0,\Mctop) -- (13.97,\Mctop);
\foreach \x in {2.08,5.08,6.50}{
  \draw[black!14, line width=0.4pt] (\x,\Mctop) -- (\x,\Mcbot);}
\node[Mcpill=cbTeam]   at (1.04,6.05) {o3-mini};
\node[Mcpill=cbMember] at (1.04,\Mcax) {Claude Sonnet 4};
\node[Mcpill=cbMoA]    at (1.04,4.75) {DeepSeek-V3};
\foreach \i in {0,...,4}{\foreach \j in {0,1,2}{
  \draw[black!52, line width=0.55pt, rounded corners=0.5pt]
    ({0.595+0.19*\i},{3.95+0.19*\j}) rectangle ++(0.13,0.13);}}
\node[Mclab, anchor=north, text width=1.65cm] at (1.04,3.85)
  {15 AIME-2024\\training problems};
\draw[Mcflow] (2.07,\Mcax) -- (2.41,\Mcax);
\draw[Mcspine,->] ($(\Mccx,\Mcax)+(82:\Mccr)$)   arc (82:-42:\Mccr);
\draw[Mcspine,->] ($(\Mccx,\Mcax)+(-58:\Mccr)$)  arc (-58:-122:\Mccr);
\draw[Mcspine,->] ($(\Mccx,\Mcax)+(222:\Mccr)$)  arc (222:98:\Mccr);
\foreach \a in {90,-50,230}{
  \filldraw[fill=white, draw=black!82, line width=1.0pt]
    ($(\Mccx,\Mcax)+(\a:\Mccr)$) circle (0.115);}
\node[Mcverb, anchor=south] at (\Mccx,6.71) {run};
\node[Mcverb, anchor=north] at (4.60,4.36) {record};
\node[Mcverb, anchor=north] at (2.62,4.36) {rewrite};
\node[Mclab, text width=1.40cm] at (\Mccx,\Mcax) {6 rounds\\per problem};
\node[Mclab, anchor=north, text width=2.85cm] at (\Mccx,3.85)
  {o3-mini rewrites the roles, phases and synthesis rules};
\draw[Mcflow] (4.86,\Mcax) -- (5.43,\Mcax);
\foreach \i in {0,...,4}{
  \draw[black!58, line width=0.6pt, rounded corners=0.5pt]
    (5.59,{4.79+0.27*\i}) rectangle ++(0.40,0.14);}
\node[Mclab, anchor=north, align=center] at (5.79,4.52)
  {5 validation\\probes};
\draw[Mcflow] (6.15,\Mcax) -- (6.73,\Mcax);
\foreach \i in {0,...,9}{
  \draw[cbTeam, fill=cbTeam!18, line width=0.55pt, rounded corners=0.6pt]
    (6.79,{4.72+0.135*\i}) rectangle ++(0.84,0.085);}
\node[Mclab, anchor=north, align=center] at (7.21,4.52) {frozen\\bank of 10};
\draw[Mcspine,->] (7.76,\Mcax) -- (8.55,\Mcax);
\draw[black!85, line width=2.4pt, line cap=butt] (\Mcwall,\Mcbot) -- ([yshift=-0.25pt]\Mcwall,\Mctop);
\McLock{\Mcwall}{\Mcax}{1.0}
\draw[Mcflow] (9.25,\Mcax) -- (12.02,\Mcax);
\fill[white] (8.57,4.86) rectangle (9.29,5.94);
\McDoc{8.63}{4.94}{0.60}{0.92}
\node[Mclab, anchor=south west, text width=1.45cm] at (8.21,6.04)
  {a new test\\problem};
\fill[white] (9.30,4.64) rectangle (10.00,6.08);
\foreach \i in {0,1}{\foreach \j in {0,...,4}{
  \draw[Mccard, rounded corners=0.7pt]
    ({9.38+0.30*\i},{4.72+0.27*\j}) rectangle ++(0.24,0.20);
  \draw[black!45, line width=0.35pt]
    ({9.42+0.30*\i},{4.85+0.27*\j}) -- ++(0.16,0);}}
\node[Mclab, anchor=north, text width=1.25cm] at (9.65,4.62) {certificates};
\node[Mcpill=cbTeam, line width=1.3pt, fill=cbTeam!20,
      font=\sffamily\footnotesize\bfseries, inner xsep=5pt, inner ysep=2.6pt]
      at (10.98,\Mcax) {o3-mini};
\node[Mclab, anchor=north, text width=1.35cm] at (10.98,4.92)
  {team member\\as judge};
\McDoc{12.15}{4.94}{0.64}{0.92}
\draw[cbGen, line width=1.6pt, line cap=round, line join=round]
  (12.30,5.30) -- (12.42,5.16) -- (12.68,5.52);
\node[Mclab, anchor=south, text width=1.2cm] at (12.47,5.98) {the answer};
\foreach \y in {6.69,6.07,5.53,4.99,4.45}{
  \draw[Mcray] (12.93,\Mcax) -- (13.90,\y);
  \fill[cbTeam!62] (13.90,\y) circle (0.055);}
\node[Mcbench, anchor=south east, align=right] at (13.94,6.86)
  {held-out AIME 2024};
\node[Mcbench, anchor=north, align=center, text width=1.35cm] at (13.42,3.85)
  {4 transfer\\benchmarks};
\end{tikzpicture}
\par\vspace{1pt}
{\centering\textbf{\small (a) Teamwork reflection and frozen deployment}\par}
\par\vspace{8pt}
\begin{minipage}{0.98\linewidth}\small
\begin{minipage}[t]{0.325\linewidth}
\begin{tcolorbox}[colback=cbTeam!3,colframe=cbTeam!60,
colbacktitle=cbTeam!13,coltitle=black,fonttitle=\rmfamily\fontsize{9}{10}\selectfont\bfseries\raggedright,
title={(i) Failure diagnosis},boxrule=0.5pt,left=3pt,right=3pt,top=2pt,bottom=2pt,
equal height group=figthreesignals,valign=top,before upper=\raggedright]
``a \textbf{regression in cross-train performance}''~\ldots\ ``\textbf{individual correct answers were lost due to team dynamics}''
\end{tcolorbox}
\end{minipage}\hfill
\begin{minipage}[t]{0.325\linewidth}
\begin{tcolorbox}[colback=cbTeam!3,colframe=cbTeam!60,
colbacktitle=cbTeam!13,coltitle=black,fonttitle=\rmfamily\fontsize{9}{10}\selectfont\bfseries\raggedright,
title={(ii) Member-specific evidence},boxrule=0.5pt,left=3pt,right=3pt,top=2pt,bottom=2pt,
equal height group=figthreesignals,valign=top,before upper=\raggedright]
``leveraging the \textbf{observed strength of Agent~2 as an auditor/challenger}''
\end{tcolorbox}
\end{minipage}\hfill
\begin{minipage}[t]{0.325\linewidth}
\begin{tcolorbox}[colback=cbTeam!3,colframe=cbTeam!60,
colbacktitle=cbTeam!13,coltitle=black,fonttitle=\rmfamily\fontsize{9}{10}\selectfont\bfseries\raggedright,
title={(iii) Strength $\rightarrow$ assigned role},boxrule=0.5pt,left=3pt,right=3pt,top=2pt,bottom=2pt,
equal height group=figthreesignals,valign=top,before upper=\raggedright]
``Agent~2, as the designated \textbf{final auditor}, must review~\ldots\ all individual attempts'' and ``\textbf{identify and re-present} any~\ldots\ claims~\ldots\ that were overlooked.''
\end{tcolorbox}
\end{minipage}
\par\vspace{5pt}
{\centering\textbf{\small (b) From failure diagnosis to a specialized agent role}\par}
\end{minipage}

\caption{\textbf{Team organization is learned offline and frozen before evaluation.} \textbf{(a)} Starting from a fixed three-model roster and $15$ AIME-2024 training problems, teamwork reflection proposes and tests organizational strategies; validation probes measure transfer, and a coverage-greedy step retains a bank of ten. At deployment, the frozen bank produces candidate certificates for a new problem, and one team member serves as judge, selecting the final answer. \textbf{(b)} In a GPQA mutation, the designated member diagnoses collaboration-induced loss and converts an observed member strength into an auditor role. Model identities were blinded during reflection; Agent~2 is Gemini-2.5-Flash. The resulting \hyperref[strat:gpqa-final-auditor]{\texttt{final\_auditor\_claim\_recovery}} strategy (Appendix~\ref{app:gpqa-strategy-bank}) solved its source problem and all five sampled validation probes and was selected into the final bank. Reflection excerpts are verbatim and lightly trimmed.}
\label{fig:evolution-pipeline}
\end{figure}

\paragraph{Teamwork reflection.}
The evolutionary search begins from an initial teamwork strategy, \(P_{\mathrm{init}}\): members first produce independent solutions, complete two rounds of debate-like exchange, and choose the final answer by majority vote over their final-round answers. Appendix~\ref{app:initialization} specifies this initialization and compares its performance with that of the learned teamwork strategies.

For each training problem $s$ we maintain an archive $H_s$ of candidate strategies and the team's executions of them. We designate the roster member with the highest training-set accuracy on the source benchmark to conduct \emph{teamwork reflection}: o3-mini for AIME-2024 and Gemini-2.5-Flash for GPQA. This member drives an evolutionary search by inspecting prior strategies, team transcripts, per-member answers, team outcomes, and validation probe results; choosing which candidate to build on; and proposing targeted mutations to roles, phases, and synthesis rules. Each proposed mutation defines a new candidate strategy, which the full team executes on the source problem; the resulting transcript and outcome return new behavioral evidence to the archive. We run six mutation rounds for each source problem, with the designated member proposing up to three candidate strategies per round. Figure~\ref{fig:evolution-pipeline}(b) illustrates one such mutation, in which the designated member converts an observed member strength into a specialized agent role.

Each mutation is developed within one source problem's archive. If it solves that source problem, we hold it fixed and evaluate it on five other training problems, which we call \emph{validation probes}. These probes measure whether the mutation transfers beyond the problem that produced it; their outcomes are written back to the archive and guide later mutations. Unlike GEPA \citep{agrawal2026gepa}, which stochastically selects a parent from an instance-wise Pareto frontier before an LM proposes a reflective mutation, our designated member chooses both which archived strategy to build on and how to mutate it after inspecting the recorded source-problem outcomes, validation probe scores, and team behavior. After teamwork reflection, we use training-set performance to greedily select and freeze a bank of up to ten complementary strategies.

\paragraph{Problem-independence.}
Two mechanisms keep problem-specific content out of the deployed strategies. During evolutionary search, a separate instance of the model used for teamwork reflection performs a semantic source-dependence audit of every field in each candidate strategy, excluding candidates that encode answer values, problem-specific facts or configurations, or source-derived solution recipes. Separately, the validation probes reward transfer: a strategy that helps only its source problem adds no cross-problem coverage and is less likely to survive coverage-greedy construction of the final strategy bank. The leakage screen guards against source-specific content, while the validation signal favors strategies whose structure transfers beyond the problem that produced them.

\paragraph{Test-time deployment.}
Given a held-out problem, we run every learned strategy to produce a pool of candidate solutions, each accompanied by a \emph{certificate}: a short, self-contained reasoning trace intended to be checkable step by step rather than a bare final answer. A single judge receives the problem and entire candidate pool in one prompt, audits every certificate for specific local defects without independently solving the problem, and selects the answer with the strongest written support (full prompts in Appendix~\ref{app:judge-prompts}). The judge is the model with the highest training-set performance on the source benchmark: o3-mini for AIME-2024 and Gemini-2.5-Flash for GPQA. We report \textbf{team coverage}, the fraction of problems whose pool contains at least one correct answer, and \textbf{team accuracy}, the fraction answered correctly after selection. The gap between them distinguishes generating a correct solution from successfully selecting it.

\section{Experimental Setup}
\label{sec:setup}
We evaluate the fixed, problem-independent strategies of Section~\ref{sec:method} on held-out source problems and transfer benchmarks. The \emph{math-and-physics suite} contains five benchmarks of competition mathematics and physics problem-solving. The \emph{knowledge-and-logic suite} contains three benchmarks spanning scientific knowledge, broad-domain question answering, and logical reasoning.

\paragraph{Splits, rosters, and deployment.}
The math-and-physics team comprises o3-mini, Claude~Sonnet~4, and DeepSeek-V3, with o3-mini conducting teamwork reflection and serving as the final judge. We learn the bank from a $15$-problem AIME-2024 training split and evaluate it on the disjoint AIME-2024 test split. We then deploy the same bank, with no further teamwork reflection or test-time controller, on four transfer benchmarks: the next two competition years (AIME~2025 and AIME~2026), a different competition (HMMT~February~2026), and a physics domain (TheoremQA-physics).

For the knowledge-and-logic suite, the team comprises Gemini-2.5-Flash, Llama-4-Maverick, and GPT-4.1, with Gemini-2.5-Flash conducting teamwork reflection and serving as the final judge. We learn a separate bank from a $25$-problem GPQA Diamond training split and evaluate it on $100$ disjoint GPQA Diamond problems. We then deploy the same bank and roster unchanged on MMLU-Pro ($n{=}100$) and five BIG-Bench Extra Hard (BBEH) logical-reasoning subtasks ($n{=}75$).

\paragraph{Baselines.}
Our primary outcome is \textbf{team accuracy}. We also report team coverage and routing-oracle coverage. The baselines target distinct alternative explanations for the team's gains. The \emph{best member} tests whether collaboration surpasses its strongest constituent. The \emph{routing oracle} is a perfect per-problem selector over the members' individual answers. Self-consistency at $K{=}10$ matches the number of candidate solutions, and five-pass self-reflection controls for additional single-agent reasoning depth. Fixed multi-agent baselines---member-vote, three-round debate, and Mixture of Agents---test whether a standard aggregation or deliberation scaffold suffices. Our strongest multi-agent control is a \emph{homogeneous team}: three copies of the member with the highest training-set accuracy (o3-mini for mathematics and physics, Gemini-2.5-Flash for knowledge and logic) execute the same frozen teamwork strategies as SAT. This holds the learned interaction structure fixed to isolate the contribution of model heterogeneity. Our strongest single-agent control is \emph{linearization}: the strongest team member, as determined by training accuracy on the source benchmark, serially carries out every role and phase in each learned strategy at approximately the team's total inference budget. Later phases receive the model's outputs from earlier phases, preserving the learned strategy structure while replacing interaction among distinct models with reasoning by a single model. An advantage over linearization can therefore reflect both multi-agent interaction and model heterogeneity. See Appendix~\ref{app:aime-stats} for evaluation details.

\section{Results}
\label{sec:main-results}
In both independent instantiations, the self-organizing team achieves the highest average accuracy among the methods we test. The math-and-physics team averages $66.7\%$ across five benchmarks, while the knowledge-and-logic team averages $72.8\%$ across three (Tables~\ref{tab:demonstrable-suite} and~\ref{tab:boundary-suite}).

\paragraph{The math-and-physics team leads in accuracy and beats the perfect routing oracle.}
Across five benchmarks, the team averages $66.7\%$ accuracy, compared with $48.8\%$ for the best member, $57.3\%$ for Mixture of Agents, $55.4\%$ for self-consistency, and $58.7\%$ for linearization (Table~\ref{tab:demonstrable-suite}). On the held-out AIME-2024 split, it reaches $84.7\%$, $20.3$ percentage points above o3-mini and above every tested single- and multi-agent control. The gap to linearization shows that o3-mini does not reproduce the team's average gain by serially replaying the same strategy structure under the matched control. The homogeneous team averages $56.0\%$ across the suite, indicating that the learned interaction structure successfully leverages model heterogeneity to improve performance. Averaged across the suite, the team's $66.7\%$ accuracy also exceeds the $59.0\%$ coverage of a perfect router over the members' individual answers. Because a perfect router can only select among those answers, surpassing this ceiling shows that the team produces correct solutions on problems no member solves independently in the observed samples.

\paragraph{AIME-trained teamwork strategies transfer across years, competitions, and domains.}
\label{sec:transfer}
Deployed unchanged on the four transfer benchmarks, the same strategies retain large gains on new competition years and distributions (Table~\ref{tab:demonstrable-suite}). On AIME~2025, the team reaches $60.8\%$, compared with $40.0\%$ for the best member and $43.3\%$ for linearization, and exceeds the routing oracle by $9.7$ percentage points. On AIME~2026, it reaches $71.2\%$, $29.0$ percentage points above the best member and $13.4$ percentage points above the routing oracle. On the harder HMMT~2026, the team and matched linearization tie at $39.4\%$, compared with $26.3\%$ for the best member. The transfer extends beyond competition mathematics: on TheoremQA-physics, the team reaches $77.2\%$, compared with $71.1\%$ for the best member and $74.0\%$ for Mixture of Agents.

\begin{table}[t]
\centering\footnotesize
\setlength{\tabcolsep}{1.6pt}
\begin{tabular}{l c c cccc c}
\toprule
& \multicolumn{1}{c}{\textbf{In-dist.}} && \multicolumn{4}{c}{\textbf{Out-of-distribution transfer}} & \\
\cmidrule(lr){2-2}\cmidrule(lr){4-7}
\textbf{Method} & AIME24 && AIME25 & AIME26 & HMMT26 & TQA-phys & \textbf{Avg} \\
\midrule
Best member                & 64.4 && 40.0 & 42.2 & 26.3 & 71.1 & 48.8 \\
\addlinespace[1pt]
Self-consistency ($K{=}10$)$^{\ddagger}$ & 75.8 && 44.4 & 53.4 & 31.9 & 71.5 & 55.4 \\
Self-reflection$^{\ddagger}$             & 71.1 && 41.1 & 52.2 & 29.3 & 72.2 & 53.2 \\
Linearization (o3-mini)$^{\ddagger}$     & 74.0 && 43.3 & 68.5 & \textbf{39.4} & 68.4 & 58.7 \\
\addlinespace[1pt]
Member-vote                 & 60.7 && 38.5 & 40.9 & 25.3 & 71.3 & 47.3 \\
Debate                      & 66.7 && 41.1 & 53.3 & 30.3 & 72.2 & 52.7 \\
Mixture of Agents           & 75.6 && 46.7 & 57.8 & 32.3 & 74.0 & 57.3 \\
Homogeneous team (o3-mini)$^{\ddagger}$ & 66.7 && 46.7 & 60.0 & 36.4 & 70.2 & 56.0 \\
\midrule
\textbf{\textsc{SAT} (Ours)} & \textbf{84.7} && \textbf{60.8} & \textbf{71.2} & \textbf{39.4} & \textbf{77.2} & \textbf{66.7} \\
\specialrule{0.35pt}{2pt}{0pt}
\rowcolor{black!5} Routing-oracle coverage  & 73.3 && 51.1 & 57.8 & 36.4 & 76.6 & 59.0 \\
\rowcolor{black!5} \textsc{SAT} coverage (Ours)  & 93.3 && 73.3 & 73.3 & 54.5 & 81.6 & 75.2 \\
\bottomrule
\end{tabular}
\caption{\textbf{\textsc{SAT} achieves the highest average accuracy across the mathematics-and-physics suite and exceeds routing-oracle coverage on average.} Strategies learned on $15$ AIME-2024 training problems are frozen and evaluated on held-out AIME-2024 problems and four transfer benchmarks. All values are percentages. Unshaded rows report final-answer accuracy; boldface marks the highest accuracy in each column, including ties. Shaded rows report perfect-selection coverage over independent member answers (routing oracle) or team certificates. Single-agent controls use the strongest team member on each benchmark unless otherwise stated; see Appendix~\ref{app:aime-stats} for evaluation details. Results report means over three seeds. $^{\ddagger}$~denotes compute controls.}
\label{tab:demonstrable-suite}
\vspace{4pt}
\end{table}

\paragraph{The knowledge-and-logic team leads in accuracy but remains below routing-oracle coverage.}
Across GPQA, MMLU-Pro, and BBEH, the self-organizing team averages $72.8\%$ accuracy, compared with $72.1\%$ for Gemini-2.5-Flash linearization, $71.2\%$ for Mixture of Agents, $70.9\%$ for the homogeneous team, $70.2\%$ for debate, and $65.9\%$ for the best member (Table~\ref{tab:boundary-suite}). SAT is best on GPQA but trails Mixture of Agents on MMLU-Pro ($82.4\%$ versus $84.3\%$). On BBEH, the homogeneous team leads at $58.7\%$, followed by debate and Mixture of Agents at $57.3\%$ and SAT at $56.0\%$. Despite these differences, SAT achieves the highest average final-answer accuracy across the suite, while remaining below the routing oracle's $79.6\%$ coverage.

\begin{table}[t]
\centering\footnotesize
\setlength{\tabcolsep}{6pt}
\begin{tabular}{lcccc}
\toprule
& \multicolumn{1}{c}{\textbf{In-dist.}} & \multicolumn{2}{c}{\textbf{Out-of-distribution transfer}} & \\
\cmidrule(lr){2-2}\cmidrule(lr){3-4}
\textbf{Method} & GPQA & MMLU-Pro & BBEH & \textbf{Avg} \\
\midrule
Best member               & 72.7 & 79.7 & 45.3 & 65.9 \\
\addlinespace[1pt]
Self-consistency ($K{=}10$)$^{\ddagger}$ & 71.6 & 81.2 & 49.8 & 67.5 \\
Self-reflection$^{\ddagger}$           & 77.7 & 81.7 & 54.7 & 71.4 \\
Linearization (Gemini-2.5-Flash)$^{\ddagger}$ & 79.0 & 84.0 & 53.3 & 72.1 \\
\addlinespace[1pt]
Member-vote               & 67.0 & 79.9 & 43.6 & 63.5 \\
Debate                    & 73.7 & 79.7 & 57.3 & 70.2 \\
Mixture of Agents         & 72.0 & \textbf{84.3} & 57.3 & 71.2 \\
Homogeneous team (Gemini-2.5-Flash)$^{\ddagger}$ & 78.0 & 76.0 & \textbf{58.7} & 70.9 \\
\midrule
\textbf{\textsc{SAT} (Ours)} & \textbf{80.0} & 82.4 & 56.0 & \textbf{72.8} \\
\specialrule{0.35pt}{2pt}{0pt}
\rowcolor{black!5} Routing-oracle coverage & 84.0 & 88.0 & 66.7 & 79.6 \\
\rowcolor{black!5} \textsc{SAT} coverage (Ours)    & 94.0 & 91.0 & 78.7 & 87.9 \\
\bottomrule
\end{tabular}
\caption{\textbf{\textsc{SAT} achieves the highest average accuracy across the knowledge-and-logic suite, but selection leaves a substantial gap to team coverage.} Across GPQA, MMLU-Pro, and BBEH, \textsc{SAT} averages $72.8\%$ final-answer accuracy, while perfect selection from its certificate pool would reach $87.9\%$. All values are percentages. Unshaded rows report final-answer accuracy; boldface marks the highest accuracy in each column, including ties. Shaded rows report perfect-selection coverage over independent member answers (routing oracle) or team certificates. Single-agent controls use the strongest team member on each benchmark unless otherwise stated; see Appendix~\ref{app:aime-stats} for evaluation details. Results report means over three seeds. $^{\ddagger}$~denotes compute controls.}
\label{tab:boundary-suite}
\end{table}

\FloatBarrier
\begin{figure}[t]
\centering
\captionsetup{skip=6pt,belowskip=0pt,singlelinecheck=false}
\captionsetup[subfigure]{skip=4pt,belowskip=0pt,singlelinecheck=true,justification=centering}
\begin{subfigure}[t]{0.49\linewidth}
  \centering
  \includegraphics[width=\linewidth,trim={0 0 233.6pt 0},clip]{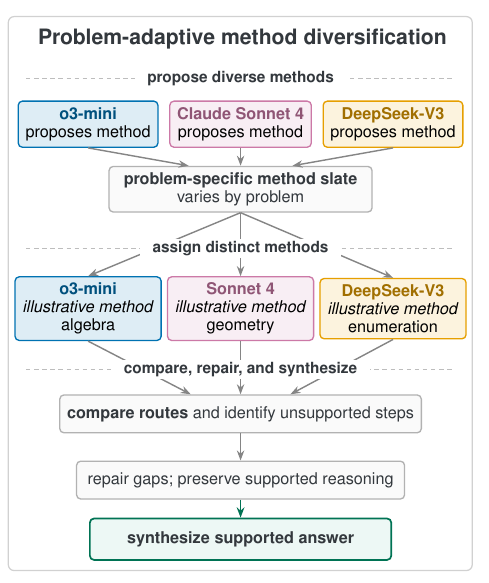}
  \caption{}\label{fig:strategy-diversification}
\end{subfigure}\hfill
\begin{subfigure}[t]{0.49\linewidth}
  \centering
  \includegraphics[width=\linewidth,trim={233.6pt 0 0 0},clip]{figures/teamwork_strategy_examples.pdf}
  \caption{}\label{fig:strategy-divergence}
\end{subfigure}
\caption{\textbf{Teamwork reflection discovers qualitatively different ways to organize reasoning.}\\
\textbf{(\subref{fig:strategy-diversification}) Problem-adaptive method diversification.} Members propose distinct approaches to the current problem, divide them among the team, and compare and repair the resulting derivations before synthesis. The strategy transfers unchanged, while the choice of methods adapts to the problem.\\
\textbf{(\subref{fig:strategy-divergence}) Divergence reconciliation.} DeepSeek-V3 can surface useful alternative derivations but also introduce errors. The strategy retains DeepSeek as a source of useful diversity while assigning o3-mini and Claude~Sonnet~4 to independently verify disputed reasoning, with DeepSeek absent from that phase. Corroborated contributions are integrated; unresolved ones are downweighted.}
\label{fig:teamwork-strategy-examples}
\end{figure}

\paragraph{Learned strategies and case studies illustrate how conversation changes the team's computation.}
The aggregate comparisons establish that strategy executions can reach answers unavailable to routing over the members' individual answers. Learned strategies enable this through reusable structures that allow flexible repair and composition of partial reasoning (Figure~\ref{fig:teamwork-strategy-examples}). For example, problem-adaptive method diversification (Figure~\ref{fig:strategy-diversification}) asks members in sequence to propose approaches suited to the current problem that differ from those already proposed. It then assigns these methods across members and compares and repairs the resulting derivations before synthesis. The strategy transfers unchanged, while the division of reasoning emerges from the problem and the members' proposals. Moreover, strategies can encode model-specific comparative advantages and behaviors. Divergence reconciliation (Figure~\ref{fig:strategy-divergence}) reflects an observed pattern in DeepSeek's behavior: its alternative derivations sometimes catch cases the other members overlook, but can also contain errors. The strategy therefore retains DeepSeek as a source of useful diversity while assigning o3-mini and Claude to independently verify its disputed reasoning, with DeepSeek absent from that phase. Additional strategy examples appear in Appendix Figure~\ref{fig:additional-strategy-examples}.

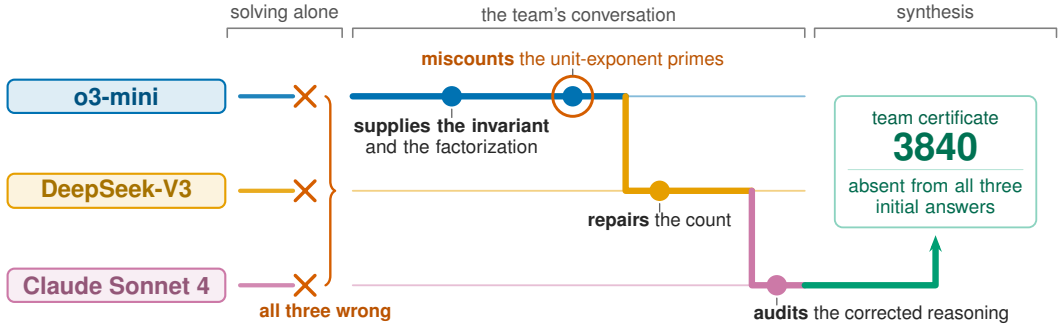
\begin{figure}[t]
\centering
\begin{tikzpicture}[
  x=1cm, y=1cm, line join=round, line cap=round,
  >={Stealth[length=2.2mm]},
  Rlypill/.style={rounded corners=2.6pt, draw=#1, fill=#1!13, text=#1!72!black,
     line width=0.8pt, font=\sffamily\footnotesize\bfseries,
     inner xsep=3pt, inner ysep=2.2pt, minimum height=0.44cm,
     minimum width=2.88cm, align=center},
  Rlyann/.style={font=\sffamily\scriptsize, text=black!85, inner sep=0pt,
     align=center, anchor=north},
  Rlyannup/.style={Rlyann, anchor=south},
  Rlyband/.style={font=\sffamily\scriptsize, text=black!68, inner sep=0pt,
     anchor=south},
  Rlylead/.style={draw=black!60, line width=0.5pt},
  Rlycard/.style={rounded corners=3pt, draw=cbGen!45, fill=white,
     line width=0.7pt, inner sep=5pt}]
\def\Rlyla{4.65}\def\Rlylb{3.40}\def\Rlylc{2.15}   %
\def\Rlybr{5.55}\def\Rlybt{5.43}                   %
\def\RlyxL{4.64}\def\RlyxR{10.62}                  %
\def\Rlyxsa{3.15}\def\Rlyxsb{3.85}\def\Rlyxx{4.01} %
\def\Rlyxbr{4.28}                                  %
\def\Rlyxa{5.95}\def\Rlyxb{7.55}                   %
\def\Rlyxda{8.25}\def\Rlyxc{8.70}\def\Rlyxdb{9.92} %
\def\Rlyxd{10.25}\def\Rlyxg{10.62}\def\Rlyxr{12.35}%
\useasboundingbox (0,1.58) rectangle (13.97,5.95);
\foreach \xa/\xb in {2.98/4.50, 4.64/10.62, 10.75/13.95}{
  \draw[black!50, line width=0.6pt]
    (\xa,\Rlybt) -- (\xa,\Rlybr) -- (\xb,\Rlybr) -- (\xb,\Rlybt);}
\node[Rlyband] at (3.74,5.63) {solving alone};
\node[Rlyband] at (7.63,5.63) {the team's conversation};
\node[Rlyband] at (12.35,5.63) {synthesis};
\node[Rlypill=cbTeam,   anchor=west] at (0.06,\Rlyla) {o3-mini};
\node[Rlypill=cbMoA,    anchor=west] at (0.06,\Rlylb) {DeepSeek-V3};
\node[Rlypill=cbMember, anchor=west] at (0.06,\Rlylc) {Claude Sonnet 4};
\foreach \c/\y in {cbTeam/\Rlyla, cbMoA/\Rlylb, cbMember/\Rlylc}{
  \draw[\c!48, line width=0.7pt] (\RlyxL,\y) -- (\RlyxR,\y);
  \draw[\c!88, line width=1.6pt] (\Rlyxsa,\y) -- (\Rlyxsb,\y);
  \draw[cbErr, line width=1.3pt]
     ({\Rlyxx-0.13},{\y-0.13}) -- ({\Rlyxx+0.13},{\y+0.13})
     ({\Rlyxx-0.13},{\y+0.13}) -- ({\Rlyxx+0.13},{\y-0.13});}
\draw[cbErr, line width=0.9pt, decorate,
      decoration={brace, amplitude=3.6pt, aspect=0.5}]
  (\Rlyxbr,\Rlyla) -- (\Rlyxbr,\Rlylc);
\node[Rlyann, text=cbErr!88!black] at (\Rlyxbr,1.91)
  {\textbf{all three wrong}};
\draw[cbTeam,   line width=2.4pt] (\RlyxL,\Rlyla) -- (\Rlyxda,\Rlyla);
\draw[cbMoA,    line width=2.4pt] (\Rlyxda,\Rlyla) -- (\Rlyxda,\Rlylb)
                                  -- (\Rlyxdb,\Rlylb);
\draw[cbMember, line width=2.4pt] (\Rlyxdb,\Rlylb) -- (\Rlyxdb,\Rlylc)
                                  -- (\Rlyxg,\Rlylc);
\fill[cbTeam]   (\Rlyxa,\Rlyla) circle (0.135);
\fill[cbTeam]   (\Rlyxb,\Rlyla) circle (0.135);
\draw[cbErr, line width=1.1pt] (\Rlyxb,\Rlyla) circle (0.265);
\fill[cbMoA]    (\Rlyxc,\Rlylb) circle (0.135);
\fill[cbMember] (\Rlyxd,\Rlylc) circle (0.135);
\draw[Rlylead] (\Rlyxa,{\Rlyla-0.17}) -- (\Rlyxa,{\Rlyla-0.28});
\node[Rlyann, text width=2.85cm] at (\Rlyxa,{\Rlyla-0.28})
  {\textbf{supplies the invariant}\\and the factorization};
\draw[Rlylead] (\Rlyxb,{\Rlyla+0.28}) -- (\Rlyxb,{\Rlyla+0.36});
\node[Rlyannup, text=cbErr!88!black]
  at (\Rlyxb,{\Rlyla+0.36}) {\textbf{miscounts} the unit-exponent primes};
\draw[Rlylead] (\Rlyxc,{\Rlylb-0.17}) -- (\Rlyxc,{\Rlylb-0.28});
\node[Rlyann] at (\Rlyxc,{\Rlylb-0.28}) {\textbf{repairs} the count};
\draw[Rlylead] (\Rlyxd,{\Rlylc-0.17}) -- (\Rlyxd,{\Rlylc-0.26});
\node[Rlyann, anchor=north west] at ({\Rlyxd-0.31},{\Rlylc-0.26})
  {\textbf{audits} the corrected reasoning};
\node[font=\sffamily\Large\bfseries, text=cbGen!72!black, inner sep=0pt]
  (Rlycb) at (\Rlyxr,3.98) {3840};
\node[font=\sffamily\scriptsize, text=cbGen!62!black, inner sep=0pt, anchor=south]
  (Rlyca) at ([yshift=0.11cm]Rlycb.north) {team certificate};
\node[font=\sffamily\scriptsize, text=cbGen!68!black, inner sep=0pt,
  align=center, text width=2.30cm, anchor=north]
  (Rlycc) at ([yshift=-0.22cm]Rlycb.south)
  {absent from all three\\initial answers};
\draw[cbGen!28, line width=0.4pt] ([yshift=0.11cm]Rlycc.north west)
  -- ([yshift=0.11cm]Rlycc.north east);
\begin{scope}[on background layer]
  \node[Rlycard, fit=(Rlyca)(Rlycb)(Rlycc)] (Rlycard) {};
\end{scope}
\draw[cbGen, line width=2.4pt, ->] (\Rlyxg,\Rlylc) -- (\Rlyxr,\Rlylc)
  -- ([yshift=-0.09cm]Rlycard.south -| \Rlyxr,0);
\end{tikzpicture}
\caption{\textbf{Cross-member repair produces an answer that no individual member initially had.} On HMMT February 2026 problem~6, all three independent answers are wrong. o3-mini supplies the invariant and factorization, DeepSeek-V3 repairs the decisive counting error, and Claude Sonnet~4 audits the correction; synthesis returns $3840$, absent from all three initial responses.}
\label{fig:conversation-score}
\end{figure}

DeepSeek's complementary value is also evident on HMMT~2026 problem~6 (Figure~\ref{fig:conversation-score}): o3-mini supplies the invariant and counting method, DeepSeek repairs a decisive counting error, and Claude audits the result. Their exchange produces the correct answer, $3840$, absent from all three initial responses (see Appendix Figure~\ref{fig:box-hmmt6} for expanded transcript). Figure~\ref{fig:shapes} contrasts these learned interaction structures with common multi-agent methods such as debate and Mixture of Agents, whose fixed protocols do not offer the same flexibility in composing and revising reasoning. Appendix Figure~\ref{fig:gpqa132-composition} provides expanded transcripts comparing the three methods on a GPQA chemistry problem. Debate turns an initially wrong majority into unanimous incorrect agreement, displacing GPT's correct answer, while Mixture of Agents propagates a mistaken identification of two reaction pathways. In contrast, \textsc{SAT} starts from three independently generated wrong answers and asks members to audit specific claims in one another's reasoning, propose corrections, and acknowledge and explain their errors. The team reaches the correct answer as a byproduct of repairing the reasoning.



Other cases show how challenges can advance an incomplete solution or protect a correct one. On AIME~2026 II-08, an integrality check exposes a gap that Claude turns into the missing construction (Figure~\ref{fig:box-ii08}). On AIME~2026 II-01, a step audit overturns an incorrect two-member majority and preserves the lone correct answer (Figure~\ref{fig:box-ii01}).

\paragraph{Team coverage and accuracy separate generation from selection.}
The math-and-physics pool averages $75.2\%$ coverage, compared with $59.0\%$ for the routing oracle; team coverage exceeds oracle coverage on all five benchmarks. The interactive strategy executions therefore yield correct certificates absent from the members' individual answers. Enough of these certificates survive selection for the team to achieve $66.7\%$ accuracy, also above the routing oracle. The knowledge-and-logic pool likewise reaches $87.9\%$ average coverage, and its per-benchmark coverage exceeds the routing oracle on all three knowledge-and-logic benchmarks. However, the team fails to convert this coverage advantage into an advantage in final-answer accuracy: its selected answers average $72.8\%$ accuracy, below the oracle's $79.6\%$ average coverage. Across benchmarks, the teams vary in how much their expanded reasoning pools translate into improvement over the strongest member. In Section~\ref{sec:demonstrability}, we propose an analytical lens for understanding why this happens.


\section{When Does Learned Organization Improve Team Performance?}
\label{sec:demonstrability}
The preceding results motivate a broader question: when can self-organizing agent teams improve over single-model performance? We use \emph{demonstrability}---an organizational-psychology construct capturing whether correct reasoning can be distinguished from incorrect reasoning \citep{laughlin1986}---as an analytical lens.

We measure demonstrability as \emph{team-certificate discriminability}. For each eligible problem, we pair one correct team certificate with one wrong team certificate, present both without correctness labels, in balanced A/B order, to a ten-model judge panel drawn from outside both deployed team rosters, and measure how often the correct certificate is selected. We average over orders, problems, and judges to obtain one benchmark-level score. The analysis includes the eight benchmarks with at least ten problems containing at least one correct and one incorrect team certificate. Unlike a formal verifier, this score does not certify an individual answer. It provides a soft, continuous benchmark-level notion of verifiability: how reliably a diverse judge panel recognizes correct reasoning relative to plausible failures.

\begin{figure}[h]
\centering
\includegraphics[width=0.58\linewidth]{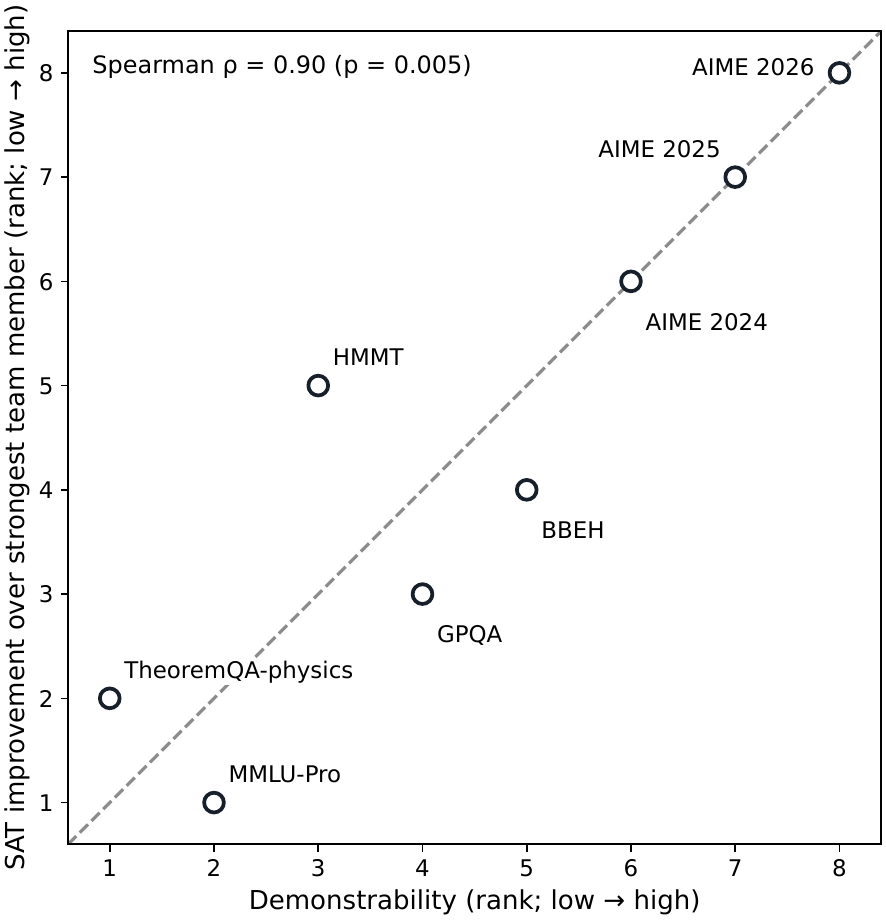}
\caption{\textbf{Demonstrability tracks when self-organizing teams improve over their strongest team member.} Across eight benchmarks, demonstrability strongly tracks the rank order of team improvement over the strongest team member (Spearman $\rho=0.90$, exact permutation $p=0.005$). Demonstrability is the balanced rate at which a ten-model panel selects correct over incorrect team reasoning. The axes rank benchmarks by demonstrability and absolute improvement over the strongest team member; the dashed diagonal denotes perfect rank agreement. Raw scores and leave-one-benchmark-out sensitivity results appear in Appendix Tables~\ref{tab:demonstrability-raw} and~\ref{tab:demonstrability-loo}.}
\label{fig:discriminability-improvement}
\end{figure}

\Needspace{10\baselineskip}
Figure~\ref{fig:discriminability-improvement} compares this benchmark-level score with absolute improvement over the best member. Across the eight benchmarks, higher demonstrability tends to coincide with a higher rank in absolute improvement over the best member. We hypothesize that this relationship reflects a simple mechanism: collaboration creates value when correct reasoning, once produced, can redirect or survive subsequent deliberation and selection, whereas correct reasoning that is difficult to recognize can be crowded out by plausible incorrect explanations. This rank association is robust to leaving out any single benchmark (Spearman $\rho=0.86$--$0.96$; see Appendix~\ref{app:demonstrability-sensitivity} for full leave-one-benchmark-out sensitivity results). We leave it to future work to test whether interventions designed to increase demonstrability within a task can improve the team's accuracy.

\FloatBarrier
\section{Related Work}
\suppressfloats[t]
\label{sec:related}
We organize prior work along three axes: \emph{how} a team computes, whether its collaboration scaffold is problem-agnostic, and whether that scaffold is learned rather than hand-designed (Table~\ref{tab:positioning}).

\paragraph{Debate and voting.}
\citet{du2024improving} introduced multi-agent debate: several model instances generate responses independently, then iteratively revise them after reading the other agents' full responses, with final answers combined by majority vote. Across seven benchmarks, \citet{choi2025debate} find that majority voting over agents' independent initial responses accounts for most of the gains attributed to debate; their theoretical analysis likewise shows that debate alone does not improve expected correctness. Across five debate methods and nine benchmarks, \citet{zhang2025stop} find that debate fails to reliably outperform single-agent chain-of-thought or self-consistency \citep[see also][]{zhu2026demystifying}. Round-table consensus methods that weight agents by confidence \citep{chen2024reconcile} share this consensus-seeking character. These findings suggest that debate often functions like voting over diverse initial responses: gains arise from sampling diversity and selection rather than agents jointly reasoning to produce new inferences.

\paragraph{Mixture of Agents and feed-forward aggregation.}
Feed-forward aggregation offers a different way to combine agent outputs: Mixture of Agents \citep{wang2025moa} layers proposer models and a synthesizer, while DyLAN \citep{liu2024dylan} stacks persona-prompted agents with inter-layer pruning. These systems can select agents, prune intermediate outputs, and synthesize across candidates, but their one-way structure does not allow agents to challenge and repair one another's reasoning through back-and-forth deliberation.

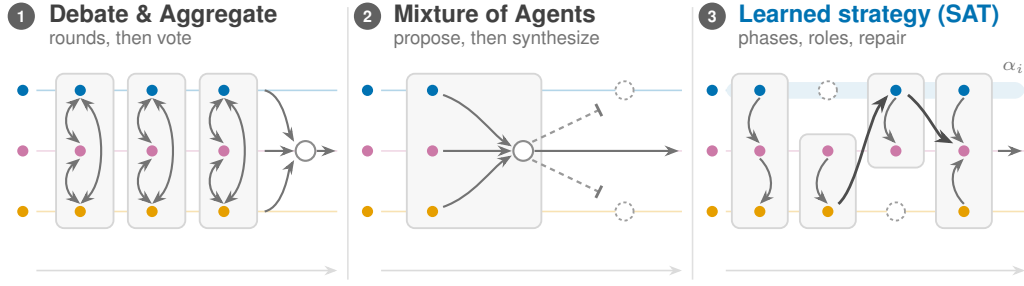
\begin{figure}[t]
\centering
\def\GshappLa{0.58}\def\GshappLb{4.40}
\def\GshappMa{5.14}\def\GshappMb{8.96}
\def\GshappRa{9.70}\def\GshappRb{13.50}
\def\GshapgOne{4.54}\def\GshapgTwo{9.10}
\def\GshapyT{3.98}\def\GshapyS{3.68}
\def\GshapyA{3.00}\def\GshapyB{2.20}\def\GshapyC{1.40}
\def\Gshapyr{0.62}
\begin{tikzpicture}[x=1cm,y=1cm,>={Stealth[length=1.9mm]},
  Gshapttl/.style={font=\footnotesize\sffamily\bfseries, inner sep=0pt,
                   anchor=west, text=black!75},
  Gshapsub/.style={font=\scriptsize\sffamily, inner sep=0pt, anchor=west,
                   text=black!55},
  Gshapnum/.style={circle, fill=black!62, text=white, inner sep=0pt,
                   minimum size=3.6mm, font=\tiny\sffamily\bfseries},
  Gshaparc/.style={black!55, line width=0.80pt, ->},
  Gshapbi/.style ={black!55, line width=0.80pt, <->},
  Gshapgone/.style={black!40, line width=0.85pt, dash pattern=on 2.0pt off 1.7pt},
  Gshapstop/.style={black!52, line width=1.10pt},
  Gshapback/.style={black!70, line width=1.05pt, -{Stealth[length=2.4mm]}},
  Gshapblk/.style={draw=black!16, line width=0.70pt, fill=black!3,
                   rounded corners=3pt},
  Gshaphub/.style={circle, draw=black!45, line width=0.80pt, fill=white,
                   inner sep=0pt, minimum size=2.6mm},
  Gshapoff/.style={circle, draw=black!38, line width=0.70pt, fill=white,
                   dash pattern=on 1.2pt off 1.1pt, inner sep=0pt,
                   minimum size=2.4mm},
  Gshaprole/.style={font=\scriptsize, inner sep=0pt, anchor=east, text=black!55}]
\useasboundingbox (0,0) rectangle (13.60,4.20);
\draw[black!12, line width=0.60pt] (\GshapgOne,0.50) -- (\GshapgOne,4.10);
\draw[black!12, line width=0.60pt] (\GshapgTwo,0.50) -- (\GshapgTwo,4.10);

\foreach \a/\b in {0.42/4.40,4.98/8.96,9.54/13.50}{%
  \draw[cbTeam!30,   line width=0.60pt] (\a,\GshapyA) -- (\b,\GshapyA);
  \draw[cbMember!30,    line width=0.60pt] (\a,\GshapyB) -- (\b,\GshapyB);
  \draw[cbMoA!30, line width=0.60pt] (\a,\GshapyC) -- (\b,\GshapyC);
  \draw[black!14, line width=0.60pt, -{Stealth[length=1.7mm]}]
    (\a,\Gshapyr) -- (\b,\Gshapyr);}
\foreach \c in {0.24,4.80,9.36}{%
  \fill[cbTeam]   (\c,\GshapyA) circle (2.1pt);
  \fill[cbMember]    (\c,\GshapyB) circle (2.1pt);
  \fill[cbMoA] (\c,\GshapyC) circle (2.1pt);}

\node[Gshapnum] at (0.22,\GshapyT) {1};
\node[Gshapttl] at (\GshappLa,\GshapyT) {Debate \& Aggregate};
\node[Gshapsub] at (\GshappLa,\GshapyS) {rounds, then vote};
\node[Gshapnum] at (4.80,\GshapyT) {2};
\node[Gshapttl] at (\GshappMa,\GshapyT) {Mixture of Agents};
\node[Gshapsub] at (\GshappMa,\GshapyS) {propose, then synthesize};
\node[Gshapnum] at (9.36,\GshapyT) {3};
\node[Gshapttl, text=cbTeam] at (\GshappRa,\GshapyT) {Learned strategy (SAT)};
\node[Gshapsub] at (\GshappRa,\GshapyS) {phases, roles, repair};

\foreach \x in {1.00,1.95,2.90}{%
  \draw[Gshapblk] (\x-0.32,1.18) rectangle (\x+0.44,3.22);
  \draw[Gshapbi] (\x,2.90) .. controls (\x-0.22,2.72) and (\x-0.22,2.48)
                           .. (\x,2.30);
  \draw[Gshapbi] (\x,2.10) .. controls (\x-0.22,1.92) and (\x-0.22,1.68)
                           .. (\x,1.50);
  \draw[Gshapbi] (\x,2.90) .. controls (\x+0.34,2.60) and (\x+0.34,1.80)
                           .. (\x,1.50);
  \fill[cbTeam]   (\x,\GshapyA) circle (2.1pt);
  \fill[cbMember]    (\x,\GshapyB) circle (2.1pt);
  \fill[cbMoA] (\x,\GshapyC) circle (2.1pt);}
\draw[Gshaparc] (3.44,3.00) .. controls (3.70,2.94) and (3.76,2.44)
                            .. (3.83,2.32);
\draw[Gshaparc] (3.44,1.40) .. controls (3.70,1.46) and (3.76,1.96)
                            .. (3.83,2.08);
\draw[Gshaparc] (3.44,2.20) -- (3.83,2.20);
\node[Gshaphub] at (3.98,\GshapyB) {};
\draw[Gshaparc] (4.13,\GshapyB) -- (4.38,\GshapyB);

\draw[Gshapblk] (5.32,1.18) rectangle (7.10,3.22);
\fill[cbTeam]   (5.66,\GshapyA) circle (2.1pt);
\fill[cbMember] (5.66,\GshapyB) circle (2.1pt);
\fill[cbMoA]    (5.66,\GshapyC) circle (2.1pt);
\draw[Gshaparc] (5.80,2.94) .. controls (6.20,2.80) and (6.45,2.42)
                            .. (6.71,2.26);
\draw[Gshaparc] (5.80,2.20) -- (6.71,2.20);
\draw[Gshaparc] (5.80,1.46) .. controls (6.20,1.60) and (6.45,1.98)
                            .. (6.71,2.14);
\node[Gshaphub] at (6.86,\GshapyB) {};
\draw[Gshaparc] (7.01,\GshapyB) -- (8.92,\GshapyB);
\draw[Gshapgone] (6.98,2.32) -- (7.86,2.74);
\draw[Gshapstop] (7.82,2.82) -- (7.90,2.66);
\draw[Gshapgone] (6.98,2.08) -- (7.86,1.66);
\draw[Gshapstop] (7.82,1.58) -- (7.90,1.74);
\node[Gshapoff] at (8.20,\GshapyA) {};
\node[Gshapoff] at (8.20,\GshapyC) {};

\fill[cbTeam!10, rounded corners=4pt] (9.54,2.89) rectangle (13.50,3.11);
\node[Gshaprole] at (13.50,3.30) {$\alpha_i$};
\draw[Gshapblk] (9.62,1.18) rectangle (10.36,3.22);
\draw[Gshapblk] (10.52,1.18) rectangle (11.26,2.42);
\draw[Gshapblk] (11.42,1.98) rectangle (12.16,3.22);
\draw[Gshapblk] (12.32,1.18) rectangle (13.06,3.22);
\draw[Gshaparc] (9.99,2.90) .. controls (9.81,2.72) and (9.81,2.48)
                            .. (9.99,2.30);
\draw[Gshaparc] (9.99,2.10) .. controls (10.17,1.92) and (10.17,1.68)
                            .. (9.99,1.50);
\fill[cbTeam]   (9.99,\GshapyA) circle (2.1pt);
\fill[cbMember]    (9.99,\GshapyB) circle (2.1pt);
\fill[cbMoA] (9.99,\GshapyC) circle (2.1pt);
\draw[Gshaparc] (10.89,2.10) .. controls (10.71,1.92) and (10.71,1.68)
                             .. (10.89,1.50);
\node[Gshapoff] at (10.89,\GshapyA) {};
\fill[cbMember]    (10.89,\GshapyB) circle (2.1pt);
\fill[cbMoA] (10.89,\GshapyC) circle (2.1pt);
\draw[Gshaparc] (11.79,2.90) .. controls (11.61,2.72) and (11.61,2.48)
                             .. (11.79,2.30);
\fill[cbTeam]   (11.79,\GshapyA) circle (2.1pt);
\fill[cbMember]    (11.79,\GshapyB) circle (2.1pt);
\node[Gshapoff] at (11.79,\GshapyC) {};
\draw[Gshaparc] (12.69,2.90) .. controls (12.51,2.72) and (12.51,2.48)
                             .. (12.69,2.30);
\draw[Gshaparc] (12.69,1.50) .. controls (12.51,1.68) and (12.51,1.92)
                             .. (12.69,2.10);
\fill[cbTeam]   (12.69,\GshapyA) circle (2.1pt);
\fill[cbMember]    (12.69,\GshapyB) circle (2.1pt);
\fill[cbMoA] (12.69,\GshapyC) circle (2.1pt);
\draw[Gshaparc] (13.14,\GshapyB) -- (13.46,\GshapyB);
\draw[Gshapback] (11.03,1.50) .. controls (11.28,1.72) and (11.42,2.62)
                              .. (11.69,2.94);
\draw[Gshapback] (11.95,2.94) .. controls (12.20,2.74) and (12.34,2.40)
                              .. (12.55,2.30);
\end{tikzpicture}
\vspace{-1.5em}
\caption{\textbf{Learned teamwork supports a richer interaction structure than common debate and feed-forward aggregation multi-agent systems.}
Debate repeats symmetric exchange before voting; Mixture of Agents aggregates independent proposals in one direction.
A learned strategy can instead vary participation across phases, maintain persistent roles, and revisit earlier reasoning for targeted repair.
For a transcript-grounded comparison on a common GPQA problem, see Appendix Figure~\ref{fig:gpqa132-composition}.}
\label{fig:shapes}
\end{figure}

Figure~\ref{fig:shapes} compares these fixed interaction structures with the participation, roles, and repair supported by \textsc{SAT}'s learned teamwork strategies.

\paragraph{Learned orchestration and topology optimization.}
A growing line optimizes multi-agent workflows and topologies. GPTSwarm \citep{zhuge2024gptswarm} represents agents as computational graphs and optimizes graph connectivity with REINFORCE; AFlow \citep{zhang2025aflow} searches code-represented workflows with Monte Carlo tree search; MASS \citep{zhou2026mass} optimizes prompts and topology over a fixed library of blocks; and AgentNet \citep{yang2025agentnet} adapts decentralized task-routing connections and per-agent retrieval memories from experience. Adjacent frameworks fix more of the surrounding program: MetaGPT \citep{hong2024metagpt} specifies a role-specialized pipeline, while DSPy \citep{khattab2024dspy} compiles prompts and demonstrations within a user-defined program. These methods optimize computational graphs, routing, or prompts around model calls; our learned artifact is instead a reusable teamwork strategy that structures a multi-phase conversation in which agents exchange and revise reasoning.

The closest comparison to our work is OPTAGENT \citep{bi2025optagent}, which uses verbal reinforcement learning to optimize the edges of a pairwise-debate graph. It learns who interacts and in what order while retaining manually specified roles, the same exchange-and-revision operation on every edge, and majority-vote selection. We instead learn reusable teamwork strategies that vary roles, participants, rounds, information flow, and synthesis, then freeze them before transfer across benchmarks.

\paragraph{Problem-conditioned task organization.}
Conductor \citep{nielsen2026conductor} trains a controller to emit a problem-conditioned workflow comprising worker instructions and communication topology. LATTE \citep{mieczkowski2026latte} instead lets agents construct and revise a shared graph of sub-tasks, dependencies, assignments, and progress during execution. Both adapt the division of labor to the current problem. We instead learn reusable teamwork strategies from training problems, freeze them before evaluation, and deploy them without prescribing a decomposition of the new problem.

Broader meta-search methods optimize agentic systems at the level of code or inference architecture. Meta Agent Search \citep{hu2025adas}, introduced as an instance of ADAS, uses a meta-agent to search open-ended agent code. Its search space can in principle express conversation, but the reported agents primarily compose model calls for generation, critique, refinement, decomposition, and integration. Archon \citep{saadfalcon2025archon} instead searches over compositions of inference-time techniques. We focus the search on reusable teamwork strategies, making the organization of multi-agent reasoning rather than a general agent program the learned artifact.

Virtual Lab \citep{swanson2025virtuallab} provides a different point of comparison: it organizes a PI-led team of domain-specialist agents through research meetings, with high-level human feedback. Its collaborative scaffold enables substantive scientific work but is not learned from prior team behavior. We instead learn reusable teamwork strategies from prior collaborations, then deploy them unchanged on held-out problems and benchmarks.

\paragraph{Open agent platforms.}
Open agent platforms provide shared environments in which many agents collaborate freely on a common problem without a prescribed workflow. EinsteinArena \citep{bianchi2026einsteinarena}, for example, allows agents to iteratively build on one another's solutions and feedback to make progress on open mathematics problems. Such platforms support collaborative computation without prescribing a problem-specific workflow, but do not learn reusable teamwork strategies from prior team behavior (Table~\ref{tab:positioning}).

\paragraph{Reflective search over reusable artifacts.}
Methodologically, we build on a line that optimizes non-weight artifacts through reflective search. GEPA \citep{agrawal2026gepa} evolves prompts by reflective mutation and retains a Pareto frontier scored per-instance on a validation split; CORAL \citep{qu2026coral} studies open-ended discovery by delegating the evolutionary process itself to autonomous agents, replacing fixed search heuristics with agent decisions over retrieval, proposal, and evaluation; and Meta-Harness \citep{lee2026metaharness} searches over harness code. Related test-time ``cheatsheet'' methods accumulate reusable, evolving context---Dynamic Cheatsheet \citep{suzgun2025dynamiccheatsheet} and Agentic Context Engineering \citep{zhang2026ace}---but these artifacts serve as memory for a single model rather than as structures for multi-agent collaboration. We apply agent-driven evolutionary search to a new object: teamwork strategies learned on per-problem archives, selected into a fixed bank using only training evidence after search, and deployed on held-out problems and benchmarks (Section~\ref{sec:method}).

\begin{table}[t]
\centering
\small
\setlength{\tabcolsep}{3pt}
\begin{tabular}{@{}lccc@{}}
\toprule
\textbf{Approach}
  & \shortstack{\textbf{Collaborative}\\\textbf{computation}\\[1pt]{\scriptsize joint reasoning, not}\\{\scriptsize aggregation or routing}}
  & \shortstack{\textbf{Problem-agnostic}\\\textbf{collaboration scaffold}\\[1pt]{\scriptsize reused across tasks}\\{\scriptsize and benchmarks}}
  & \shortstack{\textbf{Learned}\\\makebox[0pt]{\phantom{\textbf{computation}}}\\[1pt]{\scriptsize not}\\{\scriptsize hand-designed}} \\
\midrule
Debate, voting, Mixture of Agents & $\times$ & $\checkmark$ & $\times$ \\
Topology optimization (GPTSwarm, AgentNet) & $\times$ & $\times$ & $\checkmark$ \\
Interaction-graph optimization (OPTAGENT) & $\checkmark$ & $\times$ & $\checkmark$ \\
Problem-conditioned orchestration (Conductor) & $\times$ & $\times$ & $\checkmark$ \\
PI-led scientific collaboration (Virtual Lab) & $\checkmark$ & $\times$ & $\times$ \\
Open agent platform (EinsteinArena) & $\checkmark$ & $\checkmark$ & $\times$ \\
\midrule
\textbf{\textsc{SAT} (Ours)} & $\checkmark$ & $\checkmark$ & $\checkmark$ \\
\bottomrule
\end{tabular}
\caption{\textbf{A landscape of multi-agent collaboration.} Multi-agent approaches differ along three dimensions: whether agents jointly develop reasoning, whether their collaboration scaffold is agnostic to problem content and reused across tasks and benchmarks, and whether that scaffold is learned rather than hand-designed.}
\label{tab:positioning}
\end{table}

\section{Discussion and Limitations}
\label{sec:discussion}
Taken together, these experiments show that agent teams can improve by learning how to organize their reasoning. Across two domains, a designated member uses teamwork reflection on prior collaborations to construct reusable teamwork strategies that transfer unchanged to held-out problems and benchmarks. Both teams achieve the highest average accuracy among the methods tested and outperform their strongest member and matched linearization on average; in mathematics and physics, the team also surpasses a perfect router over the members' individual answers, showing that interaction can construct solutions unavailable to selection alone.

The main limitation is that richer generation does not guarantee correct selection: on knowledge and logic, the team pool reaches $87.9\%$ coverage but team accuracy reaches only $72.8\%$. Better selection or more legible certificate formats are therefore needed to turn generated solutions into final answers. Demonstrability offers a complementary but correlational analytical lens because we measure it post hoc across benchmarks and do not use it to guide teamwork reflection.

\paragraph{Future work.}
A direct next step is to optimize teamwork strategies for demonstrability by adapting roles, challenge procedures, or synthesis formats, then test whether making correct reasoning easier to distinguish improves team accuracy. Future recursive systems could also distill successful multi-agent reasoning into individual members, reassemble the strengthened team, and learn new teamwork strategies, testing whether interaction-derived capabilities survive distillation and improve the agents that drive the next cycle.

More broadly, our results position organization itself as an agent capability: without human-specified problem decompositions, a fixed set of models can learn to reason together and construct solutions unavailable to any member independently.

\enlargethispage{2\baselineskip}
\section*{Acknowledgments}
We would like to thank the members of the Zou Lab and the Kochenderfer Lab for helpful discussions and feedback.
Pappu and El gratefully acknowledge the support of the Knight-Hennessy Scholarship. 
Suzgun gratefully acknowledges the
support of a Google PhD Fellowship.
We acknowledge the use of AI tools to assist with language refinement during the writing process and code development.

\clearpage

\bibliographystyle{iclr2025_conference}
\bibliography{references}

\clearpage

\appendix
\section{Deployed strategy banks}
\label{app:strategy-banks}
This appendix lists the complete set of learned strategies deployed \emph{unchanged} at test time. Strategy instructions are transcribed verbatim from the frozen banks. Notation follows Section~\ref{sec:dsl}: each phase \textbf{P$k$} lists its participating-member set $A_k$ (the fixed roster slots \textsc{Agent 0/1/2}), its rounds $r_k$, and its information-flow mode ($\mathrm{L}$ local, $\mathrm{S}$ summary-broadcast); \emph{Roles} are the persistent per-agent prompts $\alpha$ (shown when set); the \emph{Teamwork prompt} is the shared team-level instruction $\tau$ stating collaboration norms for the whole team. In the AIME-2024 bank, Agents 0/1/2 are o3-mini, Claude~Sonnet~4, and DeepSeek-V3, respectively; in the GPQA-Diamond bank, they are Llama-4-Maverick, GPT-4.1, and Gemini-2.5-Flash, respectively. No phase, role, or teamwork prompt names a test problem. The AIME-2024 bank is also deployed unchanged across the remaining math-and-physics benchmarks (Section~\ref{sec:transfer}); the GPQA-Diamond bank is deployed unchanged on MMLU-Pro and BBEH.
{\footnotesize
\subsection{AIME-2024 strategy bank (10 strategies)}
\smallskip\par\noindent\textbf{\texttt{mechanistic\_step\_audit}} --- Agents audit the mechanics of each reasoning chain before synthesis.\par
\noindent\textbullet~\textbf{P1}~$[\{0,1,2\},\ r{=}1,\ \mathrm{L}]$: Audit the reasoning chains step by step. Each agent should identify one concrete step from another agent that is either verified, questionable, or incorrect.\par
\noindent\textbullet~\textbf{P2}~$[\{0,1,2\},\ r{=}1,\ \mathrm{L}]$: Resolve the audited issues. If a step is corrected, update the downstream calculation explicitly.\par
\noindent\textbf{Teamwork prompt.} Treat arithmetic, algebraic transformations, case splits, and counting steps as audit targets before accepting a final answer.\par
\smallskip\par\noindent\textbf{\texttt{weighted\_derivation\_consensus}} --- Agents independently recalculate disputed steps and reduce the influence of derivations that remain inconsistent.\par
\noindent\textbullet~\textbf{P1}~$[\{0,1,2\},\ r{=}1,\ \mathrm{L}]$: Each agent lists its full detailed derivation. Flag any steps that deviate from the majority computation. \textit{(per-agent --- A0: Provide your complete derivation and highlight any steps that differ from the expected methodology.; A1: Present your full computation and note any differences compared to others.; A2: Detail your derivation; if any step diverges, elaborate on your reasoning.)}\par
\noindent\textbullet~\textbf{P2}~$[\{0,1,2\},\ r{=}1,\ \mathrm{L}]$: Independently re-calculate disputed steps. Weight the contributions: assign higher influence to agents whose past derivations align with the majority (agents 0 and 1) and lower influence to agent 2 if divergence persists. \textit{(per-agent --- A0: Re-calculate and confirm your steps; compare with others to validate consistency.; A1: Verify the disputed computations, emphasizing consistency with the majority.; A2: Review your derivation in light of the group's feedback and adjust if discrepancies are found.)}\par
\noindent\textbf{Teamwork prompt.} Apply a weighted consensus mechanism during discussion: if one agent’s answer (notably agent 2) consistently deviates, reduce its weight in forming the final answer.\par
\smallskip\par\noindent\textbf{\texttt{minority\_reasoning\_challenge}} --- A designated dissenter challenges synthesis steps that may suppress correct minority reasoning before the team revises its answer.\par
\noindent\textbullet~\textbf{P1}~$[\{0,1,2\},\ r{=}1,\ \mathrm{L}]$: Each agent presents their independent reasoning, explicitly outlining their key load-bearing values and one potential source of error or dissent in the consensus derivation.\par
\noindent\textbullet~\textbf{P2}~$[\{0,1,2\},\ r{=}1,\ \mathrm{L}]$: A designated dissenting agent then challenges any step where minority correct reasoning might be suppressed, citing specific evidence from their chain.\par
\noindent\textbullet~\textbf{P3}~$[\{0,1,2\},\ r{=}1,\ \mathrm{L}]$: After addressing the challenge, the team revises and finalizes the synthesis to robustly support the final answer.\par
\noindent\textbf{Teamwork prompt.} Integrate explicit adversarial challenge steps to preserve correct minority reasoning during synthesis.\par
\smallskip\par\noindent\textbf{\texttt{divergence\_reconciliation}} --- Agents identify persistently divergent derivations, independently verify the disputed steps, and either reconcile them or reduce their influence.\par
\noindent\textbullet~\textbf{P1}~$[\{0,1,2\},\ r{=}1,\ \mathrm{L}]$: Each agent lists their full derivation. Specifically flag any steps from agents whose computations systematically differ from the majority. \textit{(per-agent --- A0: Review and note any discrepancies in the reasoning contributed by any agent showing divergence.; A1: Cross-verify computations from any agent identified with divergence using your own reasoning.; A2: If you are the agent with divergent reasoning, provide detailed explanations for each step.)}\par
\noindent\textbullet~\textbf{P2}~$[\{0,1\},\ r{=}1,\ \mathrm{L}]$: Initiate an independent re-calculation of the disputed steps provided by the identified divergent agent, comparing them with the computations from the other agents. Conclude by either reconciling the inconsistent result or reducing its influence. \textit{(per-agent --- A0: Independently re-calculate the disputed steps and share your verified results.; A1: Compare your verification results with those from the divergent input and note any differences.)}\par
\noindent\textbf{Teamwork prompt.} Introduce a reconciliation step dedicated to reviewing any persistent divergences. If an agent's contributions are repeatedly inconsistent with the group, their influence is reduced through independent verification.\par
\smallskip\par\noindent\textbf{\texttt{backward\_constraint\_validation}} --- Agents derive answer constraints, generate candidates, and verify backward.\par
\noindent\textbullet~\textbf{P1}~$[\{0,1,2\},\ r{=}1,\ \mathrm{L}]$: Derive necessary conditions for the final integer answer without relying on pre-supplied candidates: bounds, divisibility, modular residues, monotonicity, feasibility, or direct substitution conditions.\par
\noindent\textbullet~\textbf{P2}~$[\{0,1,2\},\ r{=}1,\ \mathrm{L}]$: For each candidate answer in the discussion, test it backward against the original problem conditions and the necessary conditions. Reject or repair candidates only with a stated mathematical reason.\par
\noindent\textbf{Teamwork prompt.} Use the final-answer format and problem constraints as a validation scaffold. Candidate answers must survive independent backward checks.\par
\smallskip\par\noindent\textbf{\texttt{constraint\_inventory\_then\_solve}} --- Agents first enumerate constraints and then use them to audit candidate answers.\par
\noindent\textbullet~\textbf{P1}~$[\{0,1,2\},\ r{=}1,\ \mathrm{L}]$: Do not finalize yet. List constraints any valid solution must satisfy: bounds, integrality, parity, congruences, case coverage, geometric conditions, or counting totals.\par
\noindent\textbullet~\textbf{P2}~$[\{0,1,2\},\ r{=}1,\ \mathrm{L}]$: Use the constraint inventory to check the candidate solutions. Revise only when a concrete constraint is violated or a missing case is found.\par
\noindent\textbf{Teamwork prompt.} Before accepting a final answer, build and use an explicit inventory of constraints from the problem.\par
\smallskip\par\noindent\textbf{\texttt{independent\_solve\_then\_synthesis}} --- Agents compare independent solutions, identify disagreements, and synthesize.\par
\noindent\textbullet~\textbf{P1}~$[\{0,1,2\},\ r{=}1,\ \mathrm{L}]$: Compare the independent reasoning chains. Each agent should name the answer they got, the main method they used, and one possible weakness in their own solution.\par
\noindent\textbullet~\textbf{P2}~$[\{0,1,2\},\ r{=}1,\ \mathrm{L}]$: Synthesize the strongest supported reasoning into a shared answer. If answers differ, resolve the disagreement using specific mathematical steps from the discussion.\par
\noindent\textbf{Teamwork prompt.} Preserve independent reasoning. Do not converge until each agent's solution has been compared against the others.\par
\smallskip\par\noindent\textbf{\texttt{problem\_adaptive\_method\_diversification}} --- Agents propose and pursue distinct methods suited to the current problem.\par
\noindent\textbullet~\textbf{P1}~$[\{0,1,2\},\ r{=}1,\ \mathrm{L}]$: Do not finalize yet. Each agent should propose plausible solution frames for this specific problem, such as algebraic, geometric, combinatorial, modular, invariant, constructive, extremal, coordinate, or computational-enumerative routes.\par
\noindent\textbullet~\textbf{P2}~$[\{0,1,2\},\ r{=}1,\ \mathrm{L}]$: Assign distinct viable frames across agents and push each route as far as possible. State when a route fails or supports a candidate answer. \textit{(per-agent --- A0: Prefer the first viable method family not already emphasized.; A1: Prefer a different viable method family from Agent 0.; A2: Prefer a checking or alternative method family distinct from Agents 0 and 1.)}\par
\noindent\textbullet~\textbf{P3}~$[\{0,1,2\},\ r{=}1,\ \mathrm{L}]$: Compare the method-specific results. Favor answers supported by independent routes or by the route with the clearest complete derivation.\par
\noindent\textbf{Teamwork prompt.} The team should deliberately diversify methods before synthesizing. Distinct methods are chosen based on the problem.\par
\smallskip\par\noindent\textbf{\texttt{suspicious\_consensus\_challenger}} --- If the team converges early, one agent must look for a failure mode.\par
\noindent\textbullet~\textbf{P1}~$[\{0,1,2\},\ r{=}1,\ \mathrm{L}]$: State the current consensus or disagreement. If there is a consensus, identify the weakest link in the shared reasoning.\par
\noindent\textbullet~\textbf{P2}~$[\{0,1,2\},\ r{=}1,\ \mathrm{L}]$: Agent 2 acts as consensus challenger. Try to find an alternative derivation, missing case, arithmetic error, or constraint violation. Agents 0 and 1 respond only with mathematical evidence. \textit{(per-agent --- A2: You are the consensus challenger. Look for concrete failure modes before accepting the answer.)}\par
\noindent\textbullet~\textbf{P3}~$[\{0,1,2\},\ r{=}1,\ \mathrm{L}]$: Decide whether the challenged answer survives. If it does, state why; if not, revise using the discovered issue.\par
\noindent\textbf{Teamwork prompt.} Consensus is not sufficient. If the team appears to agree, actively test whether the shared answer could still be wrong.\par
\smallskip\par\noindent\textbf{\texttt{component\_recombination\_validation}} --- Agents compute key quantities independently, reconstruct their implied components, and verify that those components recombine to satisfy the original constraints.\par
\noindent\textbf{Roles.} \textsc{Agent 0}: Share your backward validation process and adjust your computation if inconsistencies arise.; \textsc{Agent 1}: Ensure that the derived components are logically consistent and sum up to meet the given constraints.; \textsc{Agent 2}: Advocate for the correct computation if your backward validation confirms a lower and more consistent value.\par
\noindent\textbullet~\textbf{P1}~$[\{0,1,2\},\ r{=}1,\ \mathrm{L}]$: Step 1: Each agent computes the key parameter using their preferred approach and documents all critical arithmetic steps and load-bearing intermediate values. \textit{(per-agent --- A0: State your computed value along with key intermediate figures.; A1: Include a clear record of relevant arithmetic steps that support your computation.; A2: Detail your computation and list any intermediate values used in your derivation.)}\par
\noindent\textbullet~\textbf{P2}~$[\{0,1,2\},\ r{=}1,\ \mathrm{L}]$: Step 2: Perform a backward validation by using your computed value to deduce the implied component values. Verify that these components, when recombined, satisfy the overall constraints provided in the problem. Confirm that all derived components are consistent and nonnegative. \textit{(per-agent --- A0: Compute the derived components and demonstrate their recombination into the overall constraint.; A1: Verify that your deduced values jointly satisfy the aggregate condition and discuss any discrepancies.; A2: Check and confirm that all derived components are positive and consistent with the given constraints.)}\par
\noindent\textbf{Teamwork prompt.} After individual computations, collaboratively verify that the computed value leads to a consistent set of derived components, which when recombined, fully satisfy the overall constraints. Any inconsistency should trigger a review of the intermediate arithmetic for potential overestimation.\par
\subsection{GPQA-Diamond strategy bank (10 strategies)}
\label{app:gpqa-strategy-bank}
\smallskip\par\noindent\textbf{\texttt{constructive\_challenge\_and\_preservation}} --- A designated challenger proposes concrete repairs or alternatives, after which the team preserves, repairs, or rejects each well-supported claim before synthesis.\par
\noindent\textbf{Roles.} \textsc{Agent 2}: You are the primary challenger. Your role is to critically examine the claims of others, and when challenging, propose a plausible alternative interpretation, missing condition, factual error, or a specific repair to the reasoning, not just point out a flaw. Furthermore, if an individual agent presented a solution that is well-supported by evidence, you must proactively argue for its preservation, citing specific scientific principles and evidence. Your goal is to improve the scientific rigor of the solution through constructive adversarial engagement and ensure robust insights are not lost.\par
\noindent\textbullet~\textbf{P1}~$[\{0,1,2\},\ r{=}1,\ \mathrm{L}]$: Each agent independently proposes their best solution and the key scientific claims supporting it. Identify any areas of strong initial consensus or disagreement.\par
\noindent\textbullet~\textbf{P2}~$[\{0,1,2\},\ r{=}1,\ \mathrm{L}]$: Agent 2 (Challenger): Identify the weakest or most critical claim in the initial consensus or a divergent solution. Formulate a specific challenge that includes a plausible alternative interpretation, a missing condition, a factual error, or a proposed 'repair' to the reasoning. Crucially, if any agent's initial solution is well-supported by evidence, Agent 2 must articulate why that solution *should* be preserved, citing specific evidence or scientific principles. Other agents (0, 1): Defend your original reasoning with evidence or acknowledge the validity of the challenge/preservation argument and propose a repair or accept the preservation.\par
\noindent\textbullet~\textbf{P3}~$[\{0,1,2\},\ r{=}1,\ \mathrm{L}]$: Based on the challenges and responses, collectively decide whether the challenged reasoning survives, is repaired, or is abandoned. If a repair is accepted, clearly state the revised scientific claim or reasoning. If fundamental disagreements persist, especially regarding the preservation of a robust individual solution, identify the precise points of contention and their impact on the overall solution.\par
\noindent\textbullet~\textbf{P4}~$[\{0,1,2\},\ r{=}1,\ \mathrm{L}]$: Synthesize all surviving and repaired scientific claims into a final, robust solution. Clearly state the solution and justify why it is the most scientifically sound choice given the debate, explicitly mentioning how any robust individual solutions were preserved or why they were ultimately discarded.\par
\noindent\textbf{Teamwork prompt.} Engage in an adversarial debate, where challenges include proposed repairs or alternative interpretations. Critically evaluate all claims and collaboratively refine the solution, with a specific focus on preserving and validating robust individual solutions.\par
\smallskip\par\noindent\textbf{\texttt{option\_aware\_claim\_critique}} --- Agents challenge one another's scientific claims, document unresolved disagreements, and justify the final answer against the available options.\par
\noindent\textbullet~\textbf{P1}~$[\{0,1,2\},\ r{=}1,\ \mathrm{L}]$: Each agent independently analyzes the problem, proposes a solution, and identifies at least one key scientific claim or assumption in their own reasoning. Agents also propose specific checks or questions for at least one claim or assumption made by another agent, highlighting potential weaknesses or alternative interpretations.\par
\noindent\textbullet~\textbf{P2}~$[\{0,1,2\},\ r{=}1,\ \mathrm{L}]$: Agents engage in a structured debate. Each agent defends their own claim(s) with scientific evidence and addresses the proposed checks or critiques from others. Critiques should not only address scientific accuracy but also consider whether an alternative interpretation or weakness significantly impacts the choice of answer, given the available options and any other problem-specific selection constraints.\par
\noindent\textbullet~\textbf{P3}~$[\{0,1,2\},\ r{=}1,\ \mathrm{L}]$: The team synthesizes a final reasoning chain. If there are unresolved scientific disagreements, they must be explicitly stated, along with the evidence for each side. If a pragmatic decision was made to select an answer due to practical selection constraints despite scientific ambiguities, this must also be explicitly documented with justification.\par
\noindent\textbf{Teamwork prompt.} The team should engage in structured, adversarial critique, focusing on scientific claims and their implications within the problem's context. Be mindful of practical selection constraints, such as the available options, when evaluating alternative interpretations or weaknesses. The final answer must be scientifically sound and pragmatically justified.\par
\smallskip\par\noindent\textbf{\texttt{discrepancy\_and\_contradiction\_audit}} --- A designated auditor identifies contradictions and unaddressed constraints across independent solutions before the team resolves them.\par
\noindent\textbf{Roles.} \textsc{Agent 2}: You are the Discrepancy and Contradiction Auditor. Your role is to critically examine the solutions provided by Agent 0 and Agent 1. Focus on identifying and clearly articulating scientific inconsistencies, contradictions, or unaddressed problem constraints in their proposed identified entities, reactions, reasoning, and final answers. Pay close attention to details like implied properties, logical consistency, and adherence to all problem requirements. Your goal is to ensure the final team solution is scientifically robust and fully consistent with the problem statement.\par
\noindent\textbullet~\textbf{P1}~$[\{0,1,2\},\ r{=}1,\ \mathrm{L}]$: Each agent will individually present their full solution, including their final answer, detailed reasoning, identified entities, and how each explicit condition or test result stated in the problem was addressed. Do not debate yet.\par
\noindent\textbullet~\textbf{P2}~$[\{2\},\ r{=}1,\ \mathrm{L}]$: Agent 2, as the Discrepancy and Contradiction Auditor, will review all presented solutions from Agents 0 and 1. Your task is to identify and articulate any scientific inconsistencies (e.g., misinterpretations of properties, unaddressed constraints, logical flaws) AND explicit contradictions (e.g., conflicting identified entities, reaction pathways, or derivations) between the different proposed solutions or against the problem statement. Clearly present these findings to Agents 0 and 1, specifying where and why they exist.\par
\noindent\textbullet~\textbf{P3}~$[\{0,1,2\},\ r{=}1,\ \mathrm{L}]$: Agents 0 and 1 will respond to the discrepancies and contradictions raised by Agent 2. Clarify, correct, or refine your solutions based on the auditor's findings. For each identified inconsistency or contradiction, agents must engage in a structured debate to determine its root cause and decisively resolve it through scientific reasoning and reference to problem statements. Agent 2 will facilitate this discussion to ensure all issues are thoroughly addressed and resolved, leading to a unified, consistent, and scientifically robust solution.\par
\noindent\textbf{Teamwork prompt.} Independently solve the problem. Then, with a designated auditor, critically examine and resolve any scientific inconsistencies or contradictions through structured debate to arrive at a single, accurate team solution.\par
\smallskip\par\noindent\textbf{\texttt{provisional\_consensus\_cross\_validation}} --- The team forms a provisional consensus, tests every strong contender against necessary conditions, and subjects the result to a final assumptions audit.\par
\noindent\textbullet~\textbf{P1}~$[\{0,1,2\},\ r{=}1,\ \mathrm{L}]$: Each agent independently state your answer and the full scientific reasoning, including all key assumptions, critical calculations, and principles used. Share your reasoning in detail to allow for thorough cross-verification.\par
\noindent\textbullet~\textbf{P2}~$[\{0,1,2\},\ r{=}1,\ \mathrm{L}]$: Compare the independent answers and reasoning. Identify the strongest arguments and points of agreement or disagreement. Focus on any differences in assumptions or derivation steps. Propose a preliminary consensus answer based on the most robust reasoning observed among the independent solutions, explicitly noting any remaining discrepancies. If there are multiple robust and conflicting independent answers, identify them, preserving all evidence and rationale.\par
\noindent\textbullet~\textbf{P3}~$[\{0,1,2\},\ r{=}1,\ \mathrm{L}]$: For each of the preliminary consensus answer(s) (or if multiple, each strong contender), systematically cross-verify each other's full reasoning chains, including all assumptions. Identify what would need to be true if that option were correct. Include checks for consistency in units, signs, mechanism, causal direction, limiting cases, and compatibility with the experimental setup. List all derived implications and necessary conditions. Agent 0 will lead this cross-verification, focusing on quantitative consistency and scientific principles.\par
\noindent\textbullet~\textbf{P4}~$[\{0,1,2\},\ r{=}1,\ \mathrm{L}]$: Test the preliminary consensus answer(s) backward against the original question conditions and all derived implications. Systematically check each condition and implication. Reject or repair candidates only with a specific scientific reason supported by evidence. If a strong contender was not chosen as the preliminary consensus, validate it here as well. Finalize the best-supported answer based on this rigorous validation. Agent 1 will lead the final synthesis, ensuring all evidence is accounted for.\par
\noindent\textbullet~\textbf{P5}~$[\{0,1,2\},\ r{=}1,\ \mathrm{L}]$: Before finalizing, Agent 2 will perform a final audit of the chosen answer, specifically challenging any implicit assumptions and checking for alternative interpretations that might have been overlooked. The team must address these challenges before providing the final answer.\par
\noindent\textbf{Teamwork prompt.} Combine independent problem-solving with rigorous and systematic cross-verification and backward validation. Begin with independent derivations and explicit assumption declarations, synthesize a preliminary consensus by identifying strongest arguments and discrepancies, and then test this consensus and all strong contenders against the problem's conditions and all answer choices using backward reasoning and systematic checks. Conclude with a final audit of assumptions and alternative interpretations.\par
\smallskip\par\noindent\textbf{\texttt{neglected\_effects\_challenge}} --- A designated challenger quantitatively tests effects initially dismissed as negligible and requires the team to incorporate them or justify their exclusion.\par
\noindent\textbf{Roles.} \textsc{Agent 2}: You are the consensus challenger. Your primary role is to identify and quantitatively evaluate any subtle physical effects or overlooked information suggested by the problem's parameters, especially when it might change the outcome, even if initially thought to be negligible. Focus on general physical principles rather than problem-specific details.\par
\noindent\textbullet~\textbf{P1}~$[\{0,1,2\},\ r{=}1,\ \mathrm{L}]$: Each agent independently presents their initial interpretation of the problem, including relevant physical laws and initial proposed solution path. They should highlight any information they consider potentially extraneous or secondary.\par
\noindent\textbullet~\textbf{P2}~$[\{2,0,1\},\ r{=}1,\ \mathrm{L}]$: Agent 2 acts as consensus challenger. Agent 2 must scrutinize all information initially deemed 'extraneous' or 'secondary' by any agent. Specifically, Agent 2 must identify any physical phenomena (e.g., relativistic effects, quantum effects, or environmental factors) that could be implicitly suggested by the problem's parameters (e.g., high velocities, very small scales, extreme conditions) but might have been overlooked. For each identified phenomenon, Agent 2 must provide a qualitative argument for its potential relevance and an initial quantitative estimate of its impact on the solution. Agents 0 and 1 must then critically evaluate Agent 2's arguments and estimates, providing counter-arguments or supporting evidence based on general scientific principles, not on external tools or knowledge outside the prompt.\par
\noindent\textbullet~\textbf{P3}~$[\{0,1,2\},\ r{=}1,\ \mathrm{L}]$: The team collaboratively decides whether any of the challenged physical phenomena are indeed significant enough to alter the solution, based on the quantitative estimates and critical evaluations. If deemed significant, the team must explicitly incorporate the effect into their derivation and recalculate the solution. If deemed negligible, a clear, quantitative justification for its negligibility must be provided.\par
\noindent\textbullet~\textbf{P4}~$[\{0,1,2\},\ r{=}1,\ \mathrm{L}]$: Present the final, refined solution, explicitly detailing all physical effects considered, their quantitative impact (or justification for negligibility), and the final calculated answer.\par
\noindent\textbf{Teamwork prompt.} Actively challenge assumptions of negligibility, especially for physical effects suggested by problem parameters. Quantitatively evaluate potential subtle effects to determine their true significance.\par
\smallskip\par\noindent\phantomsection\label{strat:gpqa-final-auditor}\textbf{\texttt{final\_auditor\_claim\_recovery}} --- A final auditor resurfaces well-supported claims omitted from the provisional consensus for explicit integration or refutation.\par
\noindent\textbf{Roles.} \textsc{Agent 2}: You are the final auditor. Your role is critical in preventing collaboration-induced loss of well-supported individual claims. Review the provisional consensus and all individual attempts. Specifically identify and re-present any materially distinct or well-supported individual claims (especially from Agents 0 and 1) that were overlooked or not sufficiently addressed. Prompt Agents 0 and 1 to respond with supporting evidence or revised reasoning. Your goal is to ensure all valid insights are brought to the team's attention for final adjudication.\par
\noindent\textbullet~\textbf{P1}~$[\{0,1,2\},\ r{=}1,\ \mathrm{L}]$: Each agent independently derives a solution and identifies the key scientific claims, assumptions, and supporting evidence for their conclusion.\par
\noindent\textbullet~\textbf{P2}~$[\{0,1,2\},\ r{=}1,\ \mathrm{L}]$: All agents share their solutions and derivations. The team discusses to arrive at a provisional consensus solution, explicitly documenting any discrepancies or unresolved scientific issues and any unique, well-supported individual claims.\par
\noindent\textbullet~\textbf{P3}~$[\{0,1,2\},\ r{=}1,\ \mathrm{S}]$: Agent 2, as the designated final auditor, must review the provisional consensus solution and all individual attempts, paying close attention to previously documented unique or dissenting well-supported claims. If Agent 2 identifies any materially distinct or well-supported individual claims that were overlooked or not sufficiently addressed, Agent 2 must re-present the reasoning and evidence for these claims. Agents 0 and 1 respond to this audit with supporting evidence or revised reasoning. The team then collectively decides whether to integrate the overlooked claim or formally reject it with scientific justification.\par
\noindent\textbullet~\textbf{P4}~$[\{0,1,2\},\ r{=}1,\ \mathrm{L}]$: Final Adjudication: The team must now revisit ALL distinct, well-supported individual claims that were not fully integrated into the provisional consensus or were initially set aside. For each such claim, the team must explicitly discuss its scientific validity. If a claim is deemed scientifically valid based on evidence, it must be integrated into the final solution, even if it requires revising the consensus. If a claim is deemed invalid, the team must provide a clear, scientific refutation. The final team solution must represent a comprehensive adjudication of all individual insights, ensuring no well-supported individual claim is lost without thorough, explicit scientific justification.\par
\noindent\textbf{Teamwork prompt.} Independently derive, then collaboratively consolidate. A designated expert auditor will perform a final review to ensure no materially distinct or well-supported individual claims are lost. Subsequently, the team must formally adjudicate all unique individual claims, integrating valid ones and scientifically refuting invalid ones, to prevent collaboration-induced loss of valuable insights.\par
\smallskip\par\noindent\textbf{\texttt{option\_conditions\_and\_absence\_audit}} --- Members derive necessary conditions for each option and revisit eliminations that rely on the presumed absence of a signal or feature.\par
\noindent\textbullet~\textbf{P1}~$[\{0,1,2\},\ r{=}1,\ \mathrm{L}]$: Each agent independently reviews the problem and proposes an initial hypothesis for the answer, including the key supporting evidence and any assumptions made. Share these initial thoughts.\par
\noindent\textbullet~\textbf{P2}~$[\{0,1,2\},\ r{=}1,\ \mathrm{L}]$: For each answer option (A, B, C, D), Agent 0 states what scientific conditions or observations *must* be true for that option to be correct. Agent 1 then provides scientific evidence from the problem or general relevant domain principles to support or refute these conditions. Agent 2 critiques the strength of the evidence and the validity of the conditions and the interpretation.\par
\noindent\textbullet~\textbf{P3}~$[\{0,1,2\},\ r{=}1,\ \mathrm{L}]$: Based on the validation and critique, collectively eliminate options that are demonstrably false or lack sufficient evidence. Discuss any remaining ambiguities or conflicting evidence.\par
\noindent\textbullet~\textbf{P4}~$[\{0,1,2\},\ r{=}1,\ \mathrm{L}]$: Before finalizing, explicitly check if any option was eliminated based on a perceived 'absence' of a signal, property, or feature. Agent 2, specifically challenge any elimination based on the supposed 'absence' of any specific experimental observation, signal, or feature, demanding rigorous structural and mechanistic justifications for its non-existence. If so, revisit that elimination with extreme scrutiny, requiring explicit scientific reasoning for why that absence is definitive and not merely an oversight or misinterpretation.\par
\noindent\textbullet~\textbf{P5}~$[\{0,1,2\},\ r{=}1,\ \mathrm{L}]$: Synthesize the final answer from the remaining validated options, ensuring that the chosen option is fully supported by the available evidence and free of scientific contradictions.\par
\noindent\textbf{Teamwork prompt.} Systematically validate each answer option by defining its necessary conditions, evaluating evidence for those conditions, and critically reviewing the evidence. Pay special attention to claims of 'absence' of features or signals, especially in the interpretation of experimental data.\par
\smallskip\par\noindent\textbf{\texttt{corrective\_step\_audit}} --- Members audit specific scientific claims, propose and acknowledge corrections, and preserve validated minority evidence during synthesis.\par
\noindent\textbullet~\textbf{P1}~$[\{0,1,2\},\ r{=}1,\ \mathrm{L}]$: Audit the reasoning chains step by step. Each agent should identify one concrete claim from another agent that is either verified, questionable, or incorrect. For any questionable or incorrect claim, the auditing agent must explicitly propose a scientific correction and provide a brief rationale for the correction.\par
\noindent\textbullet~\textbf{P2}~$[\{0,1,2\},\ r{=}1,\ \mathrm{L}]$: Resolve the audited issues. If a claim is corrected, the agent whose claim was corrected must acknowledge the correction, explain the scientific reason for their initial error, and confirm their revised answer choice. If a proposed correction is challenged, the team must discuss and scientifically justify the most accurate correction.\par
\noindent\textbullet~\textbf{P3}~$[\{0,1,2\},\ r{=}1,\ \mathrm{L}]$: Synthesize the strongest supported answer, explicitly incorporating all agreed-upon corrections. The final answer must reflect a consolidated, scientifically sound reasoning chain, and any initial individual claims that were validated against evidence and reasoning must be explicitly identified and their preservation explained, even if initially a minority view.\par
\noindent\textbf{Teamwork prompt.} Treat scientific facts, mechanisms, definitions, unit conversions, and option eliminations as audit targets before accepting a final answer. Ensure that all identified errors are explicitly corrected and documented, and that all distinct individual claims supported by evidence are retained and integrated into the final team solution.\par
\smallskip\par\noindent\textbf{\texttt{precision\_matched\_effects\_audit}} --- The team quantifies secondary effects relative to the precision separating the answer options before selecting the closest option.\par
\noindent\textbullet~\textbf{P1}~$[\{0,1,2\},\ r{=}1,\ \mathrm{L}]$: State the current consensus or disagreement regarding the physical effects at play. If there is a consensus on which effects are relevant, identify the weakest link in the shared scientific rationale, especially regarding neglected effects or assumptions of negligibility. Explain the rationale for any initial dismissal of such effects.\par
\noindent\textbullet~\textbf{P2}~$[\{2,0,1\},\ r{=}1,\ \mathrm{L}]$: Agent 2 acts as consensus challenger. For any parameters in the problem (e.g., relative velocities, small differences in quantities, extreme conditions) that might suggest a secondary physical effect (e.g., relativistic effects, quantum effects, gravitational interactions), Agent 2 must quantitatively evaluate if this effect significantly alters the primary calculation. This evaluation must consider the magnitude of the effect relative to the precision required to distinguish between the available answer options. Agents 0 and 1 respond with their own calculations or a critical review of Agent 2's calculation, focusing on the quantitative assessment.\par
\noindent\textbullet~\textbf{P3}~$[\{0,1,2\},\ r{=}1,\ \mathrm{L}]$: The team must now perform a precise calculation of the full solution, explicitly including the quantitative impact of any physical effect identified as non-negligible. Once the numerical result is obtained, compare it against all provided multiple-choice options. For each option, calculate the absolute difference between the calculated value and the option. State the final answer as the option with the smallest absolute difference.\par
\noindent\textbullet~\textbf{P4}~$[\{0,1,2\},\ r{=}1,\ \mathrm{L}]$: Decide whether the challenged answer or current working solution survives. If it does, state why with evidence; if not, revise using the discovered issue and present the improved solution, specifically justifying the choice of the closest option based on the precise numerical comparison.\par
\noindent\textbf{Teamwork prompt.} Consensus is not sufficient. If the team appears to agree, actively test whether the shared answer could still be wrong. Specifically, always quantitatively evaluate secondary physical effects, especially when velocities, small differences, or extreme conditions are provided, to confirm their negligibility or significance, and perform precise numerical matching to options.\par
\smallskip\par\noindent\textbf{\texttt{minority\_evidence\_adjudication}} --- A designated adjudicator evaluates every claim that differs from the emerging consensus and integrates or refutes it with evidence.\par
\noindent\textbf{Roles.} \textsc{Agent 2}: You are the Minority Evidence Adjudicator. Your role is to critically evaluate any individual claims that diverge from the team's emerging consensus. You must explicitly state whether each differing claim is scientifically sound. If it is, integrate it into the collective solution and explain its relevance. If it is not, provide a clear, evidence-based scientific refutation. Be thorough and precise.\par
\noindent\textbullet~\textbf{P1}~$[\{0,1,2\},\ r{=}1,\ \mathrm{L}]$: Each agent independently derives an answer and identifies the key scientific principles, assumptions, or reasoning steps used in their derivation, and potential limitations or uncertainties in their approach.\par
\noindent\textbullet~\textbf{P2}~$[\{0,1,2\},\ r{=}1,\ \mathrm{L}]$: Identify any individual submissions that are distinct from other submissions but are scientifically sound. If such distinct submissions exist, the team must collaboratively adjudicate them, explaining why they are correct, how they relate to other submitted ideas, and whether they should be integrated or if the team's shared understanding needs to be revised. If there are no such distinct, scientifically sound submissions, simply state that all sound individual submissions align.\par
\noindent\textbullet~\textbf{P3}~$[\{0,1,2\},\ r{=}1,\ \mathrm{L}]$: Agent 2 takes on the role of the Minority Evidence Adjudicator. For every individual claim that differs from the emerging consensus, Agent 2 must explicitly state whether the claim is scientifically sound and, if so, integrate it into the collective understanding, explaining its contribution. If the claim is found to be incorrect, Agent 2 must provide a clear, scientific refutation, citing specific evidence or logical flaws. Agents 0 and 1 must engage with Agent 2's adjudication, either confirming or challenging it with further evidence. The team then synthesizes the strongest supported answer, ensuring all valid individual insights are preserved and incorrect ones are formally refuted.\par
\noindent\textbf{Teamwork prompt.} Preserve independent reasoning. Do not converge until each agent's submission and evidence have been compared against the others, and any distinct but scientifically sound individual submissions have been explicitly adjudicated, integrated, or refuted with evidence. Agent 2 is the dedicated Minority Evidence Adjudicator.\par
}

\begin{figure}[t]
\centering
\captionsetup{skip=6pt,belowskip=0pt,singlelinecheck=false}
\captionsetup[subfigure]{skip=4pt,belowskip=0pt,singlelinecheck=true,justification=centering}
\begin{subfigure}[t]{0.49\linewidth}
  \centering
  \includegraphics[width=\linewidth,trim={0 0 199pt 0},clip]{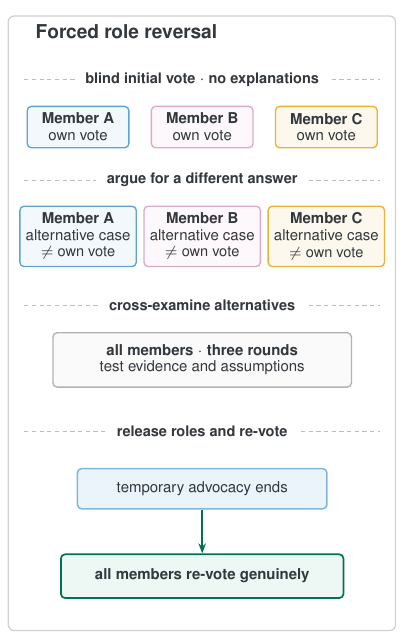}
  \caption{}\label{fig:strategy-role-reversal}
\end{subfigure}\hfill
\begin{subfigure}[t]{0.49\linewidth}
  \centering
  \includegraphics[width=\linewidth,trim={199pt 0 0 0},clip]{figures/additional_teamwork_strategy_examples.pdf}
  \caption{}\label{fig:strategy-pairwise-verification}
\end{subfigure}
\caption{\textbf{Teamwork reflection explores a broader strategy space than the final banks reveal.} \textbf{(\subref{fig:strategy-role-reversal}) Forced role reversal.} Members first vote blindly, then each argues for an answer other than its own, cross-examines the alternatives, and votes again only after the temporary advocacy roles are released. This separates the strength of an argument from its proponent's original answer and forces minority alternatives to be examined. \textbf{(\subref{fig:strategy-pairwise-verification}) Overlapping pairwise verification.} Each member states an answer and its critical assumption, after which the strategy verifies all three member pairs in an overlapping sequence. A detected flaw triggers full-team re-solving; if no flaw is found, unanimity still triggers a challenge to one key interpretation before synthesis. These strategies were discovered during earlier searches but were not selected for either final bank and were not used in reported deployment.}
\label{fig:additional-strategy-examples}
\end{figure}

Figure~\ref{fig:additional-strategy-examples} illustrates two additional strategies from earlier searches: forced role reversal and overlapping pairwise verification. Neither was selected for either deployed bank.

\section{Initialization and comparison with learned strategies}
\label{app:initialization}
The evolutionary search is initialized with \(P_{\mathrm{init}}\), in which each member first produces an independent solution. All three members then complete two rounds of debate-like exchange, each time receiving the other members' most recent responses. The protocol uses no teamwork prompt, member-specific roles, or learned phase instructions. The final answer is chosen by majority vote over the members' final-round answers; if no strict majority exists, one member's final answer is selected at random. We evaluate \(P_{\mathrm{init}}\) once per problem. Table~\ref{tab:initialization-comparison} shows higher average accuracy for \textsc{SAT} in both task suites.

\begin{table}[h]
\centering\footnotesize
\centering
\textbf{(a) Mathematics and physics}\par\smallskip
\setlength{\tabcolsep}{2.4pt}
\begin{tabular}{@{}lcccccc@{}}
\toprule
\textbf{Method} & AIME24 & AIME25 & AIME26 & HMMT26 & TQA-phys & \textbf{Avg} \\
\midrule
\(P_{\mathrm{init}}\) & 73.3 & 50.0 & 50.0 & 24.2 & 72.8 & 54.1 \\
\textbf{\textsc{SAT} (Ours)} & \textbf{84.7} & \textbf{60.8} & \textbf{71.2} & \textbf{39.4} & \textbf{77.2} & \textbf{66.7} \\
\bottomrule
\end{tabular}

\vspace{8pt}
\centering
\textbf{(b) Knowledge and logic}\par\smallskip
\setlength{\tabcolsep}{6pt}
\begin{tabular}{@{}lcccc@{}}
\toprule
\textbf{Method} & GPQA & MMLU-Pro & BBEH & \textbf{Avg} \\
\midrule
\(P_{\mathrm{init}}\) & 73.0 & 81.0 & 54.7 & 69.6 \\
\textbf{\textsc{SAT} (Ours)} & \textbf{80.0} & \textbf{82.4} & \textbf{56.0} & \textbf{72.8} \\
\bottomrule
\end{tabular}
\caption{\textbf{\textsc{SAT} achieves higher average accuracy than the initialization in both task suites.} Panels (a) and (b) compare \textsc{SAT} with \(P_{\mathrm{init}}\), the initial teamwork strategy used to seed the evolutionary search that implements teamwork reflection. The initialization has each team member propose an independent solution, followed by two rounds of debate-like exchange and a final majority vote. All values are percentages; boldface marks the higher accuracy in each column, including ties.}
\label{tab:initialization-comparison}
\end{table}

\section{Additional mechanisms of collaborative computation}
\label{app:mechanism-gallery}
\begin{figure}[t]
\centering
\includegraphics[width=\linewidth]{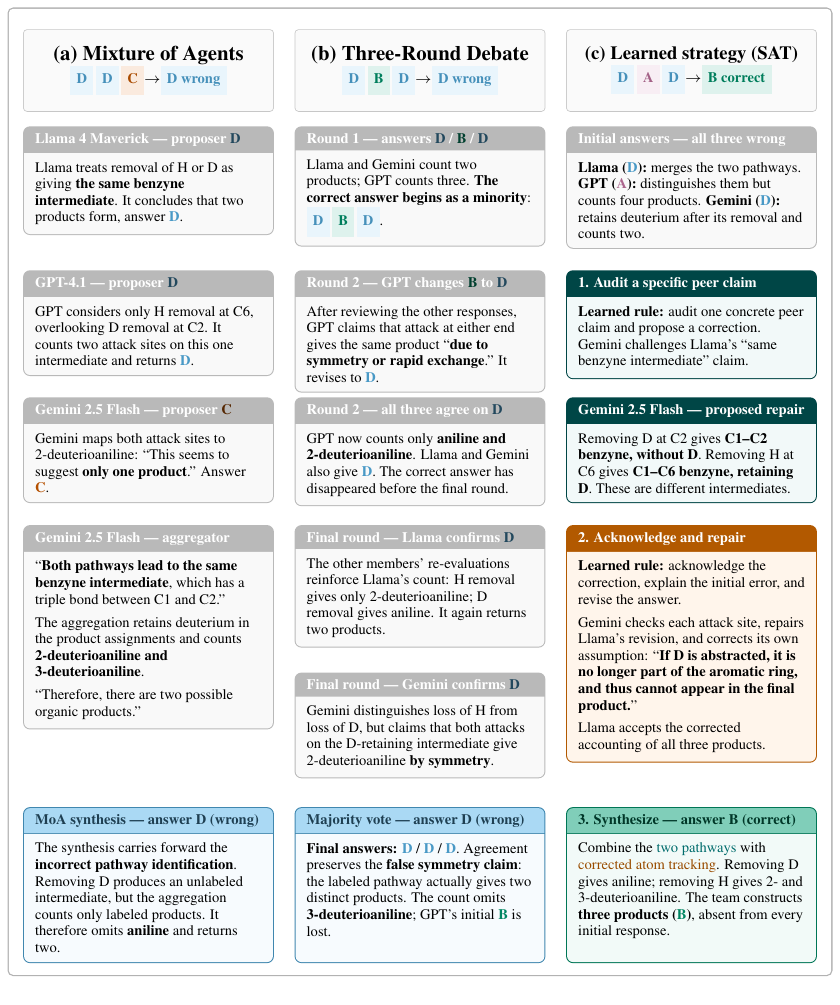}
\caption{\textbf{A learned peer-audit strategy repairs errors that debate and aggregation preserve.} GPQA Diamond~132 asks the product count for 1-bromobenzene-2-d reacting with NaNH$_2$ in ammonia. All methods use Llama~4~Maverick, GPT-4.1, and Gemini~2.5~Flash; the correct answer is B (three products). \textbf{(a) Mixture of Agents} starts with proposals D/D/C. The aggregator merges the H- and D-removal intermediates and incorrectly retains deuterium, counting 2- and 3-deuterioaniline but omitting aniline. It returns D (two). \textbf{(b) Debate} starts with a wrong D majority (Llama and Gemini); GPT initially chooses the correct answer B. Gemini's first-round analysis falsely equates the D-retaining pathway's two products. Citing the other agents' analyses, GPT adopts this equivalence and switches to D; Llama adopts the same product accounting. The wrong majority becomes unanimous, omitting 3-deuterioaniline. \textbf{(c) \textsc{SAT}} begins with all three agents independently recommending wrong answers: D/A/D (two/four/two). Its learned strategy asks each member to audit a specific peer claim, propose a correction, then acknowledge corrections, explain initial errors, and revise their answer before synthesis. Gemini challenges Llama's merged pathways, then corrects its own assumption that deuterium survives removal. The team preserves the two-pathway mechanism and repairs atom tracking: removing D gives aniline; removing H gives 2- and 3-deuterioaniline. Green marks correct B and light blue incorrect D; teal marks the audit and proposed repair, orange the acknowledged correction. Transcripts and instructions are condensed; quotations are exact.}
\label{fig:gpqa132-composition}
\end{figure}

Figures~\ref{fig:box-hmmt6} and~\ref{fig:gpqa132-composition} provide expanded transcripts for the HMMT and GPQA examples discussed in the main text. Two additional AIME cases show how a failed check motivates a missing construction (Figure~\ref{fig:box-ii08}) and how a step audit protects a correct minority (Figure~\ref{fig:box-ii01}). Transcripts are condensed for legibility.

\begin{figure}[ht]
\centering
\begin{minipage}{0.95\linewidth}\small\raggedright
\methodturn{(1)~o3-mini: invariant and counting method}{For each prime $p$, the sorted exponent list is invariant, so the final $k$-th smallest number is $\prod_p p^{E_k(p)}$. Counting by $\lfloor 2100/p^m\rfloor-\lfloor 2100/p^{m+1}\rfloor$: $E_{2026}(2)=4$, $E_{2026}(3)=3$, $E_{2026}(5)=2\ldots$ For $p=23$: $2009+88=2097$.}
\methodturn{(2)~o3-mini: factorization with one wrong count}{Final factorization $2^4\cdot3^3\cdot5^2\cdot7\cdot11\cdot13\cdot17\cdot19\cdot23$; divisor count $(4{+}1)(3{+}1)(2{+}1)(1{+}1)^{\mathbf{5}}$.}
\correctionturn{(3)~DeepSeek-V3: six unit primes, not five}{There are \textbf{6} primes $(7,11,13,17,19,23)$ with exponent $1$, not $5$. So $\tau(N)=(4{+}1)(3{+}1)(2{+}1)(1{+}1)^6=5\cdot4\cdot3\cdot64=\boxed{3840}$.}
\turn{(4)~Claude Sonnet 4: independent audit}{$p{=}2$: $131{-}65{=}66$, cumulative $2035$, and $2026\le2035$, so $E_{2026}(2){=}4$. The $p{=}3$ and $p{=}5$ counts also check; $\tau=5\times4\times3\times2^6=3840$.}
\keyturn{(5)~o3-mini: synthesis}{\methodink{$N_{2026}=2^4\cdot3^3\cdot5^2\cdot7\cdot11\cdot13\cdot17\cdot19\cdot23$}; $\methodink{\tau=(4{+}1)(3{+}1)(2{+}1)(1{+}1)^{}}\!^{\correctionink{6}}=3840$. Final answer: $\boxed{3840}$.}
\end{minipage}
\caption{\textbf{Cross-member composition constructs an answer absent from all three initial outputs.} On HMMT~2026 problem~6 (gold $3840$; all three initial answers were incorrect), o3-mini supplies the gcd--lcm exponent invariant and factorization but miscounts the unit-exponent primes. DeepSeek repairs the count, Sonnet audits it, and synthesis preserves the composed answer. The transcript is condensed for legibility.}
\label{fig:box-hmmt6}
\end{figure}

\begin{figure}[t]
\centering
\begin{minipage}{0.95\linewidth}\small
\turn{o3-mini: concrete challenge}{With $s=49$ and $b=2$, $AI=2\sqrt{49/100}=1.4$, not an integer. This is a gap. Focus on the integer condition for $AI$.}
\keyturn{Claude Sonnet 4: repair}{For $AI$ to be integer we need $b\times 7/10$ integer, so $b$ must be divisible by $10$. The smallest such $b$ is $10$, giving $s=49\times10/2=245$. The ratio is $500/24=125/6$. So $AB=245$, not $49$.}
\turn{DeepSeek-V3: minimality check}{For $b=5$, $s=122.5$ is not an integer; the least admissible $b$ is $10$, while $b=20$ gives the larger $AB=490$. Thus $245$ is minimal.}
\turn{o3-mini: completeness-gate synthesis}{Because all sides are integers, $b$ must be divisible by $10$; the least value is $b=10$, giving $s=245$ and $AI=7$. Final answer: $\boxed{245}$.}
\end{minipage}
\caption{\textbf{A failed check becomes the missing construction.} On AIME~2026 II-08 (gold $245$; initial outputs $028$/$049$/unparsed), a mandated integer check exposes the gap in the leading derivation. Sonnet turns that failed check into the divisibility constraint needed to construct $245$, and DeepSeek verifies minimality.}
\label{fig:box-ii08}
\end{figure}

\begin{figure}[t]
\centering
\begin{minipage}{0.95\linewidth}\small
\turn{o3-mini: independent answer $032$}{Considering negative common differences gives $d=-1,-2,-5,-10$. The positive cases sum to $178$ and the negative cases to $-146$, giving $\boxed{032}$.}
\keyturn{Claude Sonnet 4: step audit}{For $d=-1$, reaching $24$ requires $4+(k-1)(-1)=24$, so $k=-19$---but a term index must be positive. This eliminates every negative $d$, leaving $d\in\{1,2,5,10\}$. The answer is $178$, not $032$.}
\turn{DeepSeek-V3: revision}{Agreed---including negative $d$ required invalid negative term indices. The tenth terms for $d\in\{1,2,5,10\}$ are $13,22,49,94$, which sum to $178$.}
\end{minipage}
\caption{\textbf{Verification protects a correct minority against a wrong majority.} On AIME~2026 II-01 (gold $178$; initial answers $032$/$178$/$032$), two members share the same locally checkable error. A step audit names that error, both wrong members revise, and the team preserves the lone correct answer that voting would discard.}
\label{fig:box-ii01}
\end{figure}

\FloatBarrier
\section{Evaluation Details}
\label{app:aime-stats}
For each benchmark, self-consistency and self-reflection use the team member with the highest accuracy in the initial independent single-pass evaluation on that benchmark, breaking ties arbitrarily. Both use o3-mini for AIME~2024, AIME~2025, and AIME~2026; Claude~Sonnet~4 for TheoremQA-physics; Gemini-2.5-Flash for GPQA and BBEH; and GPT-4.1 for MMLU-Pro. On HMMT~2026, self-consistency uses Claude~Sonnet~4 and self-reflection uses o3-mini; the difference reflects arbitrary tie-breaking, as the two models had identical independent single-pass accuracy.

Self-consistency draws $K{=}10$ samples from the selected model at temperature $0.5$. HMMT~2026 answers are graded with MathArena's benchmark-specific answer checker; other mathematics answers are graded with \texttt{math-verify}. MMLU-Pro is graded by exact match to the gold option letter; BBEH uses its official evaluation logic.

\subsection{Final-answer selection prompts}
\label{app:judge-prompts}
The judge receives the benchmark problem followed by every candidate certificate in a single prompt. We use the following task-specific templates, with braced fields populated for each benchmark and problem.

\subsubsection{Free-response mathematics and physics}
\begin{lstlisting}[style=judgeprompt]
You are selecting the best proof-of-work certificate for one {benchmark_name} problem.

You are given the problem and all candidate final team certificates produced by a strategy bank. Choose the candidate whose final answer is best supported by its written certificate.

You are auditing written certificates, NOT solving the problem.

Rules:
- Do not solve the problem yourself, and do not select or reject a candidate because its answer matches or conflicts with your own solution, estimate, or intuition about the answer.
- Judge only certificate quality: are the load-bearing steps written down and locally checkable, is the arithmetic correct as written, is the case analysis complete, and does the final transition to the stated answer follow from the established steps?
- You may reject or downgrade a candidate ONLY by naming a specific local defect in its written certificate: a step that does not follow from the previous ones, an arithmetic error you can point to, a missing case, an unjustified assumption, or a final transition not supported by the steps. "I believe the answer is different" or "it conflicts with another candidate's derivation" are NOT defects.
- For each candidate, record all such defects in a named_defects list; use an empty list if no specific defect is identified.
- Do not use answer frequency while auditing; audit every certificate on its written content alone. Frequency enters ONLY through the mandatory tie-break rule in stage 2.
- If every certificate is flawed, choose the least-bad candidate and name its remaining defect.

Work in two stages, strictly in this order:
1. AUDIT: write a structured audit entry for EVERY candidate, based only on its written certificate.
2. DECIDE: only after all audits are written, compare the audits and select the certificate with the strongest written support. MANDATORY TIE-BREAK: if more than one candidate has an empty named_defects list, you MUST select from among those defect-free candidates the one whose final answer is the most common answer within that defect-free set; if frequencies are tied, prefer the certificate whose audit shows the most complete, explicit support. Do not override this tie-break with prose-quality preferences.

Problem ({benchmark_name}):
{problem}

Candidate certificates:
{candidate_certificates}
\end{lstlisting}

\subsubsection{Multiple-choice knowledge and logic}
\begin{lstlisting}[style=judgeprompt]
You are selecting the best {benchmark_name} final-answer certificate from a pool of candidate reasoning traces.

You are given one multiple-choice question and all candidate final certificates. Choose the candidate whose final answer (an option label) is best supported by its written reasoning.

You are auditing written certificates, NOT solving the problem.

Rules:
- Do not solve the problem yourself, and do not select or reject a candidate because its answer matches or conflicts with your own solution, estimate, or intuition about the correct option.
- Judge only certificate quality: are the load-bearing claims and decisive steps written down and locally checkable, are the calculations and factual claims correct as written, are the option eliminations and case distinctions complete, and does the final transition to the stated option follow from the established steps?
- You may reject or downgrade a candidate ONLY by naming a specific local defect in its written certificate: a step that does not follow from the previous ones, a calculation or factual claim you can point to as wrong as written, an unsupported option elimination, a missing case, an unjustified assumption, or a final option choice not supported by the stated reasoning. "I believe the answer is different" or "it conflicts with another candidate's reasoning" are NOT defects.
- For each candidate, record all such defects in a named_defects list; use an empty list if no specific defect is identified.
- Do not use answer frequency while auditing; audit every certificate on its written content alone. Frequency enters ONLY through the mandatory tie-break rule in stage 2.
- If every certificate is flawed, choose the least-bad candidate and name its remaining defect.

Work in two stages, strictly in this order:
1. AUDIT: write a structured audit entry for EVERY candidate, based only on its written certificate.
2. DECIDE: only after all audits are written, compare the audits and select the certificate with the strongest written support. MANDATORY TIE-BREAK: if more than one candidate has an empty named_defects list, you MUST select from among those defect-free candidates the one whose final answer is the most common answer within that defect-free set; if frequencies are tied, prefer the certificate whose audit shows the most complete, explicit support. Do not override this tie-break with prose-quality preferences.

Problem ({benchmark_name}):
{problem}

Candidate final certificates:
{candidate_certificates}
\end{lstlisting}

\FloatBarrier
\enlargethispage{2\baselineskip}
\vspace{-8pt}
\section{Demonstrability Sensitivity Analysis}
\label{app:demonstrability-sensitivity}
Table~\ref{tab:demonstrability-raw} reports the raw benchmark values underlying Figure~\ref{fig:discriminability-improvement}. The primary analysis includes benchmarks with at least ten eligible problems containing both a correct and an incorrect team certificate.
The ten panel judges are Gemma~3~4B~IT, GPT-3.5~Turbo, GPT-4.1~Nano, Command~R7B, Claude~3~Haiku, Llama~3.1~8B~Instruct, Gemma~3~12B~IT, Mistral~Small~3.2~24B~Instruct, Gemma~3n~E4B~IT, and Qwen~2.5~7B~Instruct. None is a member of either deployed team roster.

\begin{table}[h]
\centering
\small
\begin{tabular}{lrr}
\toprule
Benchmark & Demonstrability & \shortstack{Improvement over\\best member (pp)} \\
\midrule
AIME 2024 & 0.550 & 20.3 \\
AIME 2025 & 0.559 & 20.8 \\
AIME 2026 & 0.687 & 29.0 \\
HMMT 2026 & 0.519 & 13.1 \\
TheoremQA-physics & 0.423 & 6.1 \\
GPQA & 0.534 & 7.3 \\
MMLU-Pro & 0.517 & 2.7 \\
BBEH & 0.537 & 10.7 \\
\bottomrule
\end{tabular}
\caption{\textbf{Demonstrability and \textsc{SAT} improvement across eight benchmarks.} Demonstrability is operationalized as the balanced rate at which a ten-model panel selects correct over incorrect reasoning. Improvement is \textsc{SAT} accuracy minus the best member's accuracy, in percentage points.}
\label{tab:demonstrability-raw}
\end{table}

We test whether the rank association in Section~\ref{sec:demonstrability} is driven by any single benchmark by recomputing it after leaving out each benchmark in turn. Every seven-benchmark subset retains a strong positive association (Spearman $\rho=0.86$--$0.96$; exact permutation $p=0.003$--$0.024$; Table~\ref{tab:demonstrability-loo}).

\par\noindent
\begin{minipage}{\linewidth}
\centering\footnotesize
\captionsetup{font=footnotesize}
\begin{tabular}{lcc}
\toprule
Benchmark omitted & Spearman $\rho$ & Exact $p$ \\
\midrule
None (all eight) & 0.905 & 0.005 \\
\addlinespace[1pt]
AIME 2024 & 0.857 & 0.024 \\
AIME 2025 & 0.857 & 0.024 \\
AIME 2026 & 0.857 & 0.024 \\
HMMT 2026 & 0.964 & 0.003 \\
TheoremQA-physics & 0.893 & 0.012 \\
GPQA & 0.929 & 0.007 \\
MMLU-Pro & 0.893 & 0.012 \\
BBEH & 0.929 & 0.007 \\
\bottomrule
\end{tabular}
\captionof{table}{\textbf{Leave-one-out sensitivity of the demonstrability association.} The first row reports the full eight-benchmark analysis; each subsequent row reports the Spearman correlation after omitting one benchmark. Exact two-sided $p$-values enumerate all permutations of the observed outcome ranks ($8!$ for the full analysis and $7!$ for each deletion).}
\label{tab:demonstrability-loo}
\end{minipage}

\FloatBarrier

\clearpage

\end{document}